\documentclass[10pt,twocolumn,letterpaper]{article}
\usepackage[T1]{fontenc}

\usepackage{cvpr}              
\usepackage{amsmath,amssymb}
\usepackage{booktabs}
\usepackage{graphicx}
\usepackage{array}
\usepackage{multirow}
\usepackage{xcolor}
\usepackage{stfloats}
\usepackage{flafter}
\usepackage{placeins}
\usepackage{balance}
\usepackage{afterpage}
\renewcommand{\arraystretch}{1.08}
\usepackage{adjustbox}
\usepackage{booktabs}

\newcommand{\pp}{\,\mathrm{pp}}
\definecolor{resultred}{RGB}{190,32,32}
\newcommand{\best}[1]{{\color{resultred}\bfseries\boldmath #1}}
\newcommand{\second}[1]{\underline{#1}}

\newcommand{\E}{\mathbb{E}}
\newcommand{\KL}{\mathrm{KL}}
\newcommand{\CE}{\mathrm{CE}}
\newcommand{\softmax}{\operatorname{softmax}}
\newcommand{\sg}{\operatorname{sg}}

\usepackage{cuted} 

\usepackage{currfile} 

\usepackage{caption} 
\definecolor{cvprblue}{rgb}{0.21,0.49,0.74}
\usepackage[pagebackref,breaklinks,colorlinks,allcolors=cvprblue]{hyperref}

\def\paperID{*****} 
\def\confName{CVPR}
\def\confYear{2027}

\title{FedSocket: Recipient-Executable Knowledge Exchange for\\
Heterogeneous Multimodal Federated Learning}

\author{Xinyuan Zhao\\
Sun Yat-sen University}


\begin{document}
\raggedbottom
\maketitle
\begin{abstract}
Multimodal federated learning draws on complementary observations across clients, while each recipient must predict from its own available modalities for its own task. The central challenge is to make exchanged knowledge usable under these local conditions. We introduce FedSocket, a framework that jointly designs the inputs, outputs, and aggregation of a shared predictor for recipient learning and deployment. Fixed modality features provide common input coordinates independently of private encoders; task-owned components preserve local objectives; shared modality embeddings and a residual connect tasks through compatible updates. Bidirectional teaching connects private and shared learning, and clients combine the two predictors at deployment. Clients retain their native architectures and communicate predictor parameters and counts. Across six datasets, FedSocket improves missing-modality recipient accuracy over independent local training by 14.44 and 15.51 percentage points on MELD and UCF-51. It leads FML/FedKD adaptations and a monolithic shared predictor on all four primary classification and retrieval metrics in a common protocol. At matched inference capacity, it outperforms independent ensembles by 11.06 points in UCF-51 pooled accuracy and 4.87 points in mean bidirectional Flickr30k recall at rank one (R@1). FedSocket turns complementary supervision and cross-task knowledge into stronger predictions that recipients can deploy.

\end{abstract}

\begin{figure*}[t]
\centering
\includegraphics[width=\textwidth]{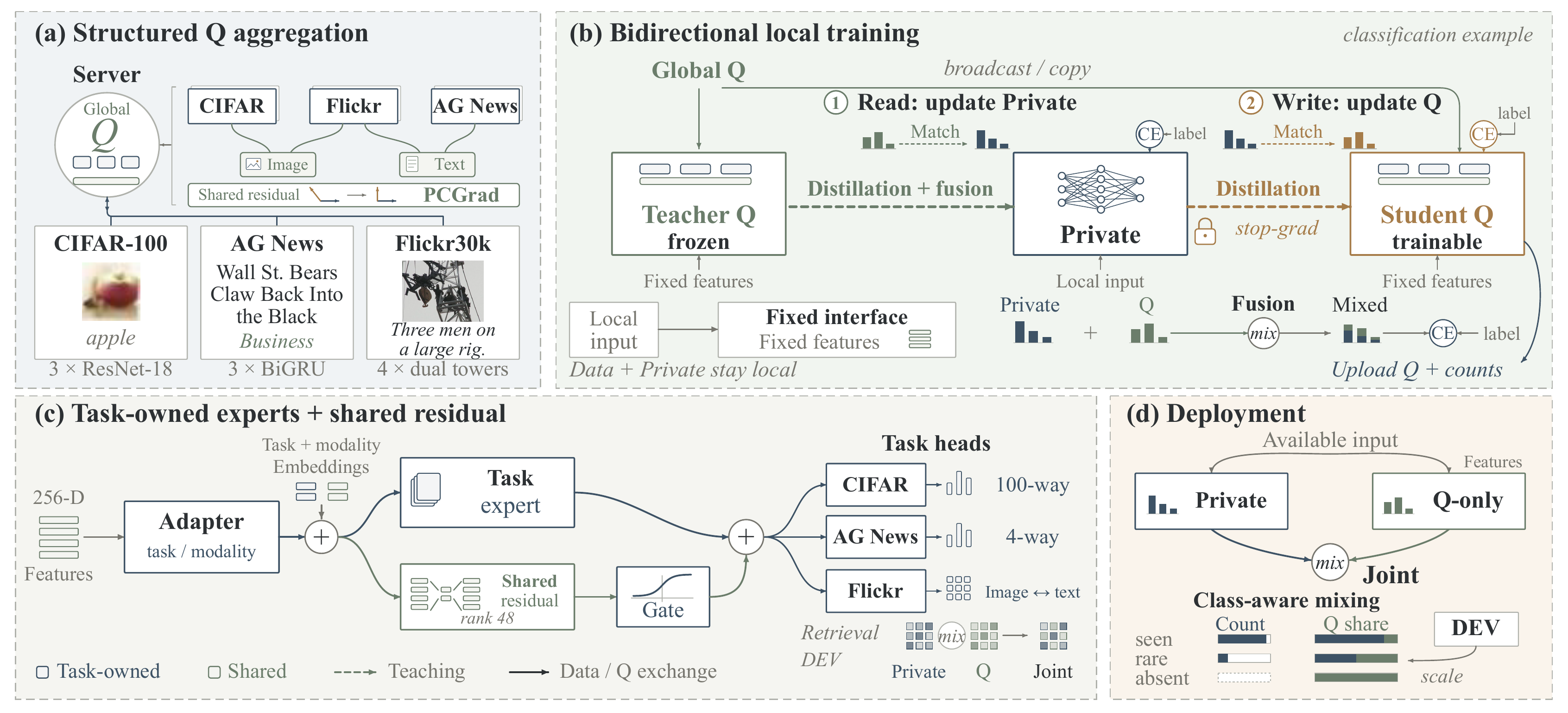}
\caption{FedSocket in the task-heterogeneous setting. Q is the shared predictor; Private is the client-owned model.
(a) Task-owned blocks aggregate within tasks; Image/Text denote shared modality embeddings, with curved links identifying the tasks that use them. PCGrad (projecting conflicting gradients) projects conflicting shared-residual updates before averaging. Uploads contain Q parameters and sample/class counts.
(b) At round $r$, Global Q ($Q^r$) initializes frozen Teacher Q ($Q^T$) and trainable Student Q ($Q^S$). Read updates Private from Teacher Q; Write trains Student Q from Private. Distillation matches predictions; training fusion uses labels to supervise their mixture. ``Stop-grad'' means stop-gradient, keeping Private fixed during Write. CE is cross-entropy; bars show class probabilities. Private uses its native inputs, while Q reads fixed features.
(c) Adapters transform inputs; task experts learn task-specific features; the shared residual adds cross-task features, scaled by a learned gate. Task heads produce predictions; matrices show retrieval scores. 256-D means 256-dimensional features, rank 48 specifies residual rank, and 100-way/4-way denote class counts.
(d) Q-only runs aggregated Q; Joint mixes Q and Private outputs. DEV (development data) selects inference weights. Count is the local class count; Q share is Q's prediction weight, which increases for rare or absent classes.
ResNet-18 is an 18-layer residual network; BiGRU is a bidirectional gated recurrent unit; dual towers are separate image/text encoders. $3\times/4\times$ indicate client counts. CIFAR/Flickr abbreviate the CIFAR-100/Flickr30k datasets; AG News is the news-topic dataset. ``St.'' in the news example abbreviates Street.}
\label{fig:method}
\end{figure*}

\section{Introduction}
Multimodal federated learning aims to improve local predictions using complementary observations distributed across clients while keeping data local. The opportunity comes from their differences: some clients observe richer modalities, others provide knowledge from related tasks. These clients also use models suited to their own resources and objectives~\cite{feng2023fedmultimodal,ouyang2023harmony,yu2023creamfl}. To benefit from collaboration, a receiving client must turn this distributed knowledge into predictions using the inputs it actually observes. This makes recipient prediction the organizing objective of knowledge exchange.

Three requirements must hold together. First, a recipient missing a modality needs a predictor whose inputs remain locally computable. Second, different private encoders require a common exchange interface that stays stable as their representations change. Third, different prediction objectives require task-specific outputs and a rule for sharing the parameters that support them. These requirements connect local execution to server aggregation: the federation must learn knowledge through components that recipients can use. They arise both when clients solve one task with unequal modality access and when different tasks learn from overlapping modalities.

Shared-model federation provides a foundation. FML connects personalized models through mutual learning and supports different local tasks through partial-model sharing~\cite{shen2020fml}. FedKD exchanges a small mentee model, using adaptive mutual distillation and compressed updates to reduce communication~\cite{wu2022fedkd}. FedSocket addresses how to construct and aggregate the exchanged predictor when modality access and prediction spaces vary together. Our insight is to tie its input interface and task outputs to the ownership of its parameters, so that the same exchange structure supports both local learning and recipient deployment.

We introduce \textbf{FedSocket} for \emph{recipient-executable knowledge exchange}. Its shared predictor Q reads fixed modality features independently of private encoders. Task-owned adapters, experts, and heads preserve each prediction objective, while shared modality embeddings and a gated residual carry knowledge across tasks. The server follows this structure: task-owned components merge within tasks, modality embeddings merge across users of the same modality, and residual updates are coordinated across tasks. This design links what each recipient can execute to what the federation aggregates.

Q makes aggregated knowledge available during both training and inference. Private models teach Q, and the returned Q supervises private learning. At deployment, Joint prediction combines Q with the private model, retaining access to federated knowledge alongside local expertise. Clients keep their native architectures and exchange Q parameters and counts. We evaluate this deployment role against independent ensembles at matched capacity.

Experiments on six datasets establish these benefits at the recipient. Joint prediction improves missing-modality recipient accuracy over independent local training by $14.44$ and $15.51$ percentage points (pp) on MELD and UCF-51. FedSocket leads FML/FedKD adaptations and Naive Shared on all four primary classification and retrieval metrics under a common protocol. At matched inference capacity, it outperforms independent ensembles by $11.06\pp$ in UCF-51 pooled accuracy and $4.87\pp$ in mean bidirectional Flickr30k retrieval recall at rank one (R@1). Teacher and task-isolated controls establish the benefits of complete-modality supervision and cross-task sharing beyond within-task federation; training controls quantify local coordination.

Our contributions are:
\begin{itemize}
  \item \textbf{Recipient-executable exchange.} We formulate recipient-executable exchange by coupling locally computable inputs, task-valid outputs, and parameter ownership under modality, model, and task heterogeneity.
  \item \textbf{Task-structured prediction and aggregation.} FedSocket implements this principle with fixed modality interfaces, task-owned components, shared modality embeddings, and a gated residual. Corresponding aggregation rules and bidirectional teaching connect Q to heterogeneous private models.
  \item \textbf{Evidence for stronger recipient deployment.} Six-dataset experiments establish gains over local training, FML/FedKD adaptations, and independent ensembles at matched capacity. Teacher and task-isolated controls establish the benefits of complete-modality supervision and cross-task knowledge exchange.
\end{itemize}

\section{Related Work}
\paragraph{Heterogeneous federated knowledge exchange.}
FedAvg aggregates a shared model, while FedProx regularizes local optimization under statistical and systems heterogeneity~\cite{mcmahan2017fedavg,li2020fedprox}. Knowledge distillation transfers predictions between models with different architectures~\cite{hinton2015distillation}. FedMD exchanges predictions on common public examples~\cite{li2019fedmd}; FedGEMS selects client knowledge for a larger server model, and FedET uses heterogeneous ensembles~\cite{cheng2021fedgems,cho2022fedet}. FedProto exchanges class prototypes to regularize local models~\cite{tan2022fedproto}. The exchanged predictions or representations impose different requirements on client inputs and output spaces.

FML mutually trains personalized and meme models; a shared partial model can be combined with a local adapter for different tasks~\cite{shen2020fml}. FedKD couples local mentors with a shared mentee through adaptive distillation and compresses its updates for communication efficiency~\cite{wu2022fedkd}. MH-pFLID uses a public-data-free messenger with receiver and transmitter modules for heterogeneous private architectures~\cite{xie2024mhpflid}.

FedSocket focuses on the structure of the exchanged predictor under coupled modality and task differences. Its fixed modality mappings decouple Q's inputs from private encoder states; task-owned components preserve output meaning; modality embeddings and a shared residual define cross-task exchange, with aggregation following parameter ownership. This structure makes Q usable for both local supervision and Joint prediction. \Cref{tab:capability} compares exchange requirements. The direct FML/FedKD adaptations also receive fixed interfaces, task-valid outputs, and Joint inference, providing a common setting for comparing shared-model designs (\cref{tab:neighbor-main}).

\paragraph{Multimodal federation and missing modalities.}
FedMultimodal benchmarks modality heterogeneity~\cite{feng2023fedmultimodal}. FedMSplit supports multimodal multi-task collaboration with heterogeneous active sensors, and Harmony disentangles multimodal training across clients with different modalities~\cite{chen2022fedmsplit,ouyang2023harmony}. MFedMC and BMSFed select clients or modality components to control communication and modality bias~\cite{yuan2024mfedmc,fan2024bmsfed}. Federated prompt tuning aligns instructions for heterogeneous and incomplete multimodal inputs~\cite{phung2025fedprompt}; ModalityMirror and TACTFL transfer or align information when modalities are missing~\cite{feng2025modalitymirror,sun2025tactfl}. FedSocket addresses how clients with different private architectures and tasks can train, aggregate, and run the same exchanged predictor.

\paragraph{Cross-task representation exchange.}
CreamFL uses public image--text representations to support exchange across models, modalities, and tasks~\cite{yu2023creamfl}. FedHTCM combines cross-modal distillation with task-conflict handling~\cite{yin2025fedhtcm}. FedAMB addresses modality dominance through adaptive distillation~\cite{han2026fedamb}, and FedAFD combines alignment, adversarial fusion, and distillation~\cite{tan2026fedafd}. FedSocket combines task-owned outputs with compatible shared Q parameters. Task-owned heads preserve prediction objectives; modality embeddings and a shared residual carry cross-task knowledge. Task-isolated Q measures the overall benefit of this exchange beyond within-task federation.

\paragraph{Transfer objectives and multimodal optimization.}
Contrastive representation distillation and cross-modal KD improve knowledge transfer between representations or modalities~\cite{tian2020crd,chen2023amid}. OGM and PMR address imbalanced multimodal optimization~\cite{peng2022ogm,fan2023pmr}. Our conditional target follows Bregman prediction~\cite{banerjee2005conditional}, and compatible shared updates use PCGrad-style projection~\cite{yu2020pcgrad}.

\section{Recipient-Executable Knowledge Exchange}
Let $\mathcal K$ be the set of clients. Client $k\in\mathcal K$ owns data $\mathcal D_k=\{(x_i^{\mathcal M_k},y_i)\}$ and private model $f_k(\cdot;\theta_k)$: $i$ indexes examples, $x_i^{\mathcal M_k}$ contains the inputs from its available modality set $\mathcal M_k$, $y_i$ is the task target, and $\theta_k$ contains its private parameters. Clients differ in modality access, private architecture, and task output space. We denote the exchanged predictor by Q. The participating clients are registered during training; Q serves each client through its available modalities and task output space.

We instantiate the same exchange principle in two settings. Within one task, let $S$ denote the recipient's observed modality set and $U$ the additional donor modalities. Donors observe $(x_S,x_U)$ and teach Q using complete-modality predictions; recipients run Q through $x_S$. Across tasks, clients share compatible Q components while retaining outputs supervised by their own task. Donor and recipient describe knowledge-flow roles: in cross-task collaboration, every client receives global Q and teaches a local Q copy.

A shared predictor becomes executable when its input coordinates, supervised outputs, and parameter ownership agree with the recipient's interface. For modality $m$, a fixed mapping $h_m$ converts its input $x^m$ into features $u^m=h_m(x^m)$ for $Q(\cdot;\phi)$, where $\phi$ contains Q's parameters. Task and modality identifiers accompany these features. Q implements three requirements (\cref{fig:method}).
\paragraph{Input computability.} Every Q input uses an available modality independently of private encoder states. Clients execute the same $h_m$ for modality $m$ while retaining their native private architectures.
\paragraph{Output validity.} Each task supervises its own classification or retrieval outputs; local distillation compares predictions within that task's output space.
\paragraph{Aggregation ownership.} The server merges corresponding Q blocks according to their task or modality ownership. Residuals carry cross-task updates; heads are averaged only within their task.
\label{sec:interface-contract}
For example, classification and retrieval clients can share the embedding for their image modality and the residual module while retaining separate prediction heads. The fixed mappings keep Q's inputs stable throughout local training; the sharing rules determine which clients update each parameter block.

\section{FedSocket: Learning and Deploying Q}
\subsection{Recipient-Computable Q and Task-Owned Outputs}
Q reads fixed features computed from locally available modalities. These features provide common input coordinates across clients, while each private model learns its own representation. In cross-task experiments, frozen ResNet-18 image features and GloVe-mean text features are projected into the 256-dimensional real vector space $\mathbb R^{256}$.

For task $t$ and modality $m$, the adapter $A_{t,m}$ transforms input feature $u=u^m$ into Q's hidden space. Learned task and modality embeddings $e_t,e_m$ are vectors that identify the input's task and modality. Together they form $b=A_{t,m}(u)+e_t+e_m$. A task expert $E_t$ learns task-specific transformations, and shared residual module $R$ adds information learned across tasks:
\begin{equation}
v_{t,m}(u)=E_t(b)+\sigma(g_t)R(b).
\label{eq:qhidden}
\end{equation}
Here $v_{t,m}$ is the hidden feature and $g_t$ is a learned task scalar. Sigmoid $\sigma(a)=1/(1+e^{-a})$ sets how much the residual contributes. Hidden width is 256 and residual rank is 48. Classification heads map hidden features to class logits; the retrieval head produces normalized embeddings for matching inputs. Adapters, experts, and heads specialize within tasks, while $R$ and modality embeddings connect tasks through shared updates.

For missing-modality transfer, a donor supervises Q with predictions from its complete inputs, while Q reads the recipient-observed inputs. Let $t_k(x_S,x_U)$ be donor $k$'s temperature-scaled class probabilities and $z=h_S(x_S)$ the fixed observed-modality features. Under donor population $D$, the ideal target is
\begin{equation}
Q_S^*(z)=\mathbb E_D[t_k(x_S,x_U)\mid h_S(x_S)=z].
\label{eq:conditional-target}
\end{equation}
The conditional expectation averages teacher predictions over donor examples with the same observed features; the star denotes the optimal population predictor. It minimizes forward Kullback--Leibler (KL) risk through this interface~\cite{banerjee2005conditional}. This projection motivates Q supervision from complete-modality predictions; label anchoring and reliability weighting shape the local training objective. Thus, complete-modality supervision enters through training while recipient execution uses the available inputs. \Cref{sec:projection-proof} gives the projection and label-anchoring analysis.

\subsection{Bidirectional Local Knowledge Transfer}
Each round lets the private model read aggregated knowledge and write its local knowledge back to Q (\cref{fig:method}(b)). After private warm-up, round $r$ broadcasts global predictor $Q^r=Q(\cdot;\phi^r)$. Each client makes two copies: frozen teacher $Q^T$ and trainable student $Q^S$. The teacher first supervises Private; the updated Private then teaches the student through detached predictions. The superscripts $T$ and $S$ denote these training roles. For classification, let $p$ be Private logits, $q^T,q^S$ the teacher/student logits, and $y$ the labels:
\begin{align}
\mathcal L_{\mathrm{priv}}&=\CE_{\mathrm{ls}}(p,y)
+\lambda_{\mathrm{KD}}\mathcal L_{\mathrm{RKD}}(p,\sg(q^T);y)
\nonumber\\[-1pt]
&\quad+\lambda_{\mathrm{fus}}\mathcal L_{\mathrm{fus}}(p,q^T,y),\\
\mathcal L_Q&=\CE_{\mathrm{ls}}(q^S,y)
+\lambda_{\mathrm{rel}}\mathcal L_{\mathrm{RKD}}(q^S,\sg(p);y).
\label{eq:local-objectives}
\end{align}
The objectives $\mathcal L_{\mathrm{priv}}$ and $\mathcal L_Q$ update Private and the student Q, respectively. Label-smoothed cross-entropy $\CE_{\mathrm{ls}}$ supplies label supervision. Reliability-weighted knowledge distillation (KD), $\mathcal L_{\mathrm{RKD}}$, teaches one predictor to match another's reliable soft predictions; its second argument is the teacher and $\sg$ stops gradients. Fusion loss $\mathcal L_{\mathrm{fus}}$ uses labels to train the combined Private/teacher-Q prediction. Its learned client mixing weight $g_k^{\mathrm{train}}$ controls their probability mixture (\cref{eq:classification-fusion-training}). The nonnegative coefficients $\lambda_{\mathrm{KD}},\lambda_{\mathrm{fus}},\lambda_{\mathrm{rel}}$ weight returned KD, fusion, and Private-to-Q teaching. Retrieval applies these roles to image--text score distributions and multi-positive matching. MELD uses group cross-fitting, with each target group excluded from teacher training.

\subsection{Compatible Server Aggregation}
The server combines clients' student-Q updates according to which tasks and modalities use each block (\cref{fig:method}(a)). Clients upload Q parameters and sample/class counts. Task-owned adapters, experts, task embeddings, gates, and heads are averaged within their task; each classification-head row uses clients with positive counts for that class. A modality embedding receives updates from tasks using that modality, and the shared residual receives updates from all tasks.

For task set $\mathcal T$, let $\Delta_t$ be task $t$'s mean residual-parameter update. Projecting Conflicting Gradients (PCGrad)-style projection removes the component opposed to another task's update when their inner product is negative~\cite{yu2020pcgrad}. With projected updates $\widetilde\Delta_t$, the server uses
\begin{equation}
\phi_R^{r+1}=\phi_R^r+|\mathcal T|^{-1}\sum_t\widetilde\Delta_t.
\end{equation}
Here $\phi_R^r$ contains residual parameters at round $r$, and $|\mathcal T|$ counts tasks. Image embeddings connect CIFAR/Flickr; text embeddings connect AG/Flickr. The same-task instance uses sample-weighted Q averaging and a 0.5 server stabilizer. Aggregation details appear in \cref{sec:method-details}.

\subsection{Reusing Q at Recipient Inference}
The same recipient inputs support three deployment modes (\cref{fig:method}(d)): Private runs $f_k$; Q runs the aggregated predictor through fixed modality features; Joint combines their outputs. For task-heterogeneous classification, Q receives more weight for classes with fewer local training examples and full weight for absent classes. Development data (DEV) select the strength of this class-count rule. UCF selects a scalar Q weight on recipient DEV, and retrieval selects a score mixture for each direction. These inference weights are separate from Q's residual gate and the training-time fusion weight. Checkpoints and coefficients are fixed before test evaluation (TEST); calibration is detailed in \cref{sec:method-details}.

\section{Experimental Design}
\subsection{Protocol Instances}
We evaluate recipient-executable exchange across six datasets. MELD, CREMA-D, and UCF-51 test complete-modality supervision for recipients that lack audio. CIFAR-100, AG News, and Flickr30k test task-structured exchange across classification, retrieval, and heterogeneous private architectures. Same-task clients use sample-weighted Q averaging; heterogeneous tasks separately aggregate task-owned blocks and shared residuals. ``Public Q'' denotes the globally shared predictor. Main results and primary ablations use five paired training seeds.

Endpoint and capacity comparisons test recipient gains and the value of deploying Q. Teacher controls isolate complete-modality supervision; task-isolated Q measures the overall contribution of cross-task sharing; KD$\times$fusion tests local coordination; backbone groups examine architectural diversity.

\Cref{tab:protocol-main} summarizes the interfaces; full protocols appear in \cref{tab:protocol}. Accuracy (Acc.) measures correct classifications; macro F1 averages classwise F1; unweighted average recall (UAR) averages classwise recall. Retrieval directions I2T/T2I mean image-to-text/text-to-image. R@1/5/10 measures queries with a correct match among the top 1/5/10 candidates. Flickr30k uses five-fold 200-image evaluation, aggregated once per seed. Complete endpoints appear in \cref{tab:recipient-endpoints}.

\begin{table}[t]
\centering\footnotesize\setlength{\tabcolsep}{3pt}
\caption{Evaluation settings. T/A/V/I: text/audio/video/image. Arrows indicate donor-to-recipient modality access. Flickr uses five-fold 200-image evaluation. Full protocols: \cref{tab:protocol}.}
\label{tab:protocol-main}
\begin{tabular*}{\columnwidth}{@{\extracolsep{\fill}}lll@{}}
\toprule
Dataset & Client inputs & Primary metric \\
\midrule
\multicolumn{3}{@{}l}{\emph{Same task with missing modalities}} \\
MELD & T+A $\rightarrow$ T & Accuracy \\
CREMA-D & V+A $\rightarrow$ V & Accuracy \\
UCF-51 & V+A $\rightarrow$ V & Accuracy \\
\midrule
\multicolumn{3}{@{}l}{\emph{Heterogeneous tasks and architectures}} \\
CIFAR-100 & 3 image clients & Accuracy \\
AG News & 3 text clients & Accuracy \\
Flickr30k & 4 image--text clients & I2T/T2I R@1 \\
\bottomrule
\end{tabular*}
\end{table}

\begin{table}[t]
\caption{Same-task Joint accuracy (\%; mean $\pm$ SD, five seeds). All pools client test samples; Rec. contains missing-modality recipients. Red/underline: best/second-best means.}
\label{tab:same-task-main}
\centering\footnotesize\setlength{\tabcolsep}{4pt}
\begin{tabular*}{\columnwidth}{@{\extracolsep{\fill}}llcc@{}}
\toprule
Dataset & Method & All & Rec. \\
\midrule
\multirow{4}{*}{MELD} & Local & $54.13\pm0.95$ & $53.81\pm6.54$ \\
 & Label Q & \second{$61.90\pm0.78$} & $66.97\pm2.21$ \\
 & Single-modal Q & $61.60\pm0.69$ & \second{$67.23\pm2.01$} \\
 & FedSocket & \best{$62.16\pm0.77$} & \best{$68.24\pm1.99$} \\
\midrule
\multirow{4}{*}{CREMA-D} & Local & $74.79\pm0.32$ & \second{$74.65\pm0.50$} \\
 & Label Q & \second{$74.81\pm0.59$} & $74.48\pm0.64$ \\
 & Single-modal Q & $74.49\pm0.38$ & $74.12\pm0.64$ \\
 & FedSocket & \best{$75.45\pm0.34$} & \best{$75.24\pm0.46$} \\
\midrule
\multirow{4}{*}{UCF-51} & Local & $71.70\pm0.13$ & $71.32\pm0.21$ \\
 & Label Q & $81.22\pm0.78$ & $84.62\pm1.12$ \\
 & Single-modal Q & \second{$81.73\pm0.22$} & \second{$85.32\pm0.35$} \\
 & FedSocket & \best{$82.83\pm0.93$} & \best{$86.83\pm1.31$} \\
\bottomrule
\end{tabular*}
\end{table}

\subsection{Statistical and Selection Protocol}
Results report means and sample standard deviations (SDs) over five training seeds on fixed primary partitions. Full/control contrasts pair the same training seeds (\cref{tab:mechanism-absolute}). Paired effects include unadjusted 95\% Student-$t$ confidence intervals (CIs) and exact two-sided sign-flip $p$-values; \cref{sec:paired-summary} defines the statistics. Same-task All accuracy pools test samples, heterogeneous classification averages fixed clients, and Flickr aggregates five retrieval folds within each seed.

DEV selects checkpoints and fusion coefficients before TEST. Private, Q, and Joint use the same test examples within each run. All, Recipient, and Backbone group denote pooled clients, missing-modality recipients, and clients with a specified private architecture, respectively.

\subsection{Baselines, Controls, and Ablations}
\paragraph{Knowledge sources.} Full denotes FedSocket with all training and sharing components enabled; Local trains each private model independently. Label Q retains label training and removes Private$\rightarrow$Q supervision. Single-modal Q uses a text-only teacher on MELD or a video-only teacher on CREMA-D/UCF, keeping donor Private multimodal and recipient Q inputs fixed. Task-isolated Q preserves training and aggregation within each task and disables cross-task exchange; its comparison with Full measures the overall sharing benefit.
\paragraph{Training and deployment.} No return disables Q$\rightarrow$Private KD and fusion training. Their factorial toggles the two terms separately while retaining Private$\rightarrow$Q teaching. Private, Q, and Joint distinguish local prediction, shared prediction, and their combination. Independent predictors and checkpoint ensembles test matched-capacity deployment (\cref{sec:ensemble-controls}).
\paragraph{Shared-model baselines.} FML/FedKD adaptations and Naive Shared receive the same fixed input interfaces, task output spaces, and Joint selection rules. These comparisons evaluate FedSocket's task-structured exchange against alternative shared-model designs at matched parameter budgets (\cref{sec:neighbor-controls}). Published reference results appear in \cref{tab:reference-heterogeneous}.

\section{Results}
\subsection{Recipient Gains with Missing Modalities}
\begin{figure}[t]
\centering
\includegraphics[width=0.82\linewidth]{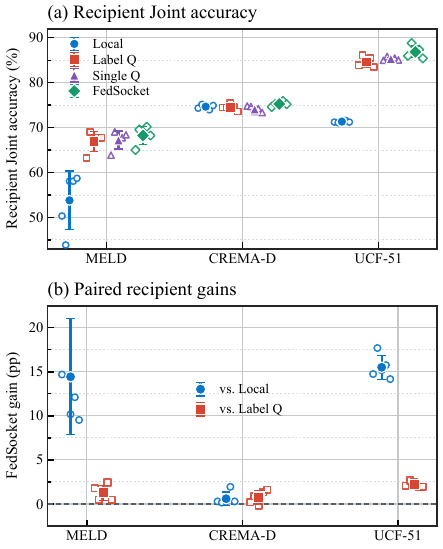}
\caption{Returned Q improves missing-modality recipient prediction. (a) Joint accuracy; (b) paired gains over Local and Label Q, measuring complete-system gains and the contribution of teacher supervision. Small points: seeds; large points: means; error bars: sample SD.}
\label{fig:core-endpoints}\label{fig:same-task}\label{fig:population}
\end{figure}

FedSocket turns exchanged knowledge into stronger prediction at recipients that lack audio (\cref{tab:same-task-main,fig:same-task}). Recipient Joint accuracy reaches $68.24\%$ on MELD and $86.83\%$ on UCF-51, improving over Local by $14.44\pm6.59\pp$ and $15.51\pm1.35\pp$. Retaining Q at deployment adds $14.04\pp$ and $14.56\pp$ beyond Private on these datasets. On CREMA-D, FedSocket achieves the highest mean recipient Joint accuracy in \cref{tab:same-task-main} at $75.24\%$. The recipient-computable predictor is therefore a direct source of deployment gains. \Cref{tab:recipient-endpoints} reports complete class-balanced metrics and the paired training/deployment decomposition for all three datasets.

Complete-modality teacher supervision further strengthens the exchanged predictor. Full exceeds Label Q in recipient Joint accuracy by $1.27\pm0.84\pp$, $0.76\pm0.76\pp$, and $2.22\pm0.35\pp$ on MELD, CREMA-D, and UCF-51. The paired differences are positive in all five MELD/UCF seeds and four CREMA-D seeds. Full also exceeds Single-modal Q by $1.01\pp$ on MELD and $1.51\pp$ on UCF-51. These controls keep recipient inputs fixed while varying the teacher's information: audio-informed supervision strengthens prediction without requiring audio at the recipient.

\subsection{Joint Prediction Across Classification and Retrieval}
\begin{table}[t]
\caption{Task-heterogeneous prediction (\%; mean $\pm$ sample SD, five seeds). Classification reports accuracy and macro F1 together; Flickr reports five-fold 200-image R@1. Red/underline: best/second-best means.}
\label{tab:heterogeneous-main}\label{tab:classification-main}
\centering\small\setlength{\tabcolsep}{2pt}

\begin{tabular}{@{}lccc@{}}
\toprule
Task / metric & Private & Public Q & Joint \\
\midrule
CIFAR Acc. & $45.13\pm0.39$ & \second{$50.32\pm0.48$} & \best{$53.59\pm1.28$} \\
CIFAR F1 & $36.88\pm0.43$ & \second{$49.44\pm0.48$} & \best{$49.70\pm1.23$} \\
AG News Acc. & $49.73\pm0.48$ & \second{$75.85\pm1.11$} & \best{$78.50\pm0.42$} \\
AG News F1 & $38.79\pm0.65$ & \second{$74.25\pm1.57$} & \best{$76.90\pm0.42$} \\
\midrule
Flickr I2T R@1 & $35.09\pm0.26$ & \second{$37.86\pm0.82$} & \best{$42.37\pm0.51$} \\
Flickr T2I R@1 & $30.38\pm0.45$ & \second{$31.13\pm0.73$} & \best{$35.46\pm0.48$} \\
\bottomrule
\end{tabular}

\end{table}

Joint improves over both Private and Q on all six classification and retrieval metrics (\cref{tab:heterogeneous-main}). Paired gains over Q are $3.27\pm1.68\pp$ on CIFAR, $2.65\pm0.90\pp$ on AG, and $4.50\pm0.41\pp$/$4.33\pm0.82\pp$ on Flickr I2T/T2I, positive in every seed (\cref{tab:mechanism-contrasts}). Joint also leads at all reported retrieval ranks (\cref{tab:recipient-endpoints}), establishing its deployment value across distinct prediction spaces.

\begin{figure}[t]
\centering
\includegraphics[width=\linewidth]{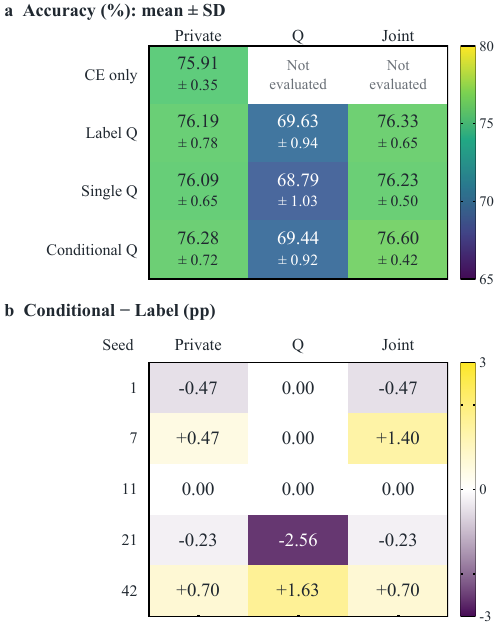}
\caption{Prediction partners in a two-stage CREMA-D development study (D03; 430 audit examples). (a) Accuracy (\%), mean $\pm$ sample SD over five seeds. CE only trains Private from labels; Label, Single, and Conditional Q use labels, video-only teachers, and full-modality teachers, respectively. Private/Q/Joint deploy the local predictor/shared predictor/their mixture. Unevaluated endpoints are marked. (b) Paired Conditional-minus-Label differences (pp), with seeds as rows: purple/white/yellow denote negative/zero/positive values. Panels use separate color scales. Selection and audit protocol: \cref{sec:development-diagnostics}.}
\label{fig:diag-d03}
\end{figure}

\paragraph{Direct shared-model comparisons.}
FedSocket leads FML-adapted, FedKD-adapted, and Naive Shared on all four mean Joint metrics (\cref{tab:neighbor-main}). The methods share partitions, private backbones, recipient interfaces, 40 rounds, and Joint selection rules; controls have 913,052 shared parameters versus Q's 913,003. FedSocket gains $12.42\pp$/$3.87\pp$ over FML-adapted and $13.49\pp$/$8.23\pp$ over FedKD-adapted on CIFAR/AG. Gains over Naive Shared reach $14.24\pp$/$8.36\pp$ on CIFAR/AG and $1.29\pp$/$0.98\pp$ on Flickr I2T/T2I. The matched budgets expose the benefit of task-structured prediction and aggregation. Full endpoints, costs, and adaptations appear in \cref{tab:neighbor-endpoints,sec:neighbor-controls}.

\begin{table}[t]
\centering\footnotesize\setlength{\tabcolsep}{1.5pt}
\caption{Direct shared-model controls at Joint inference (\%; mean $\pm$ SD, five seeds). $^*$FML/FedKD are protocol adaptations; Naive denotes Naive Shared. CIFAR/AG: accuracy; Flickr: five-fold R@1. Endpoint, paired, and resource details: \cref{tab:neighbor-endpoints}.}
\label{tab:neighbor-main}
\begingroup\renewcommand{\best}[1]{{\color{resultred}#1}}
\begin{tabular*}{\columnwidth}{@{\extracolsep{\fill}}lcccc@{}}
\toprule
Method & CIFAR & AG & I2T & T2I \\
\midrule
\textbf{FedSocket} & \best{$53.59\!\pm\!1.28$} & \best{$78.50\!\pm\!0.42$} & \best{$42.37\!\pm\!0.51$} & \best{$35.46\!\pm\!0.48$} \\
FML$^*$ & $41.17\!\pm\!0.49$ & $74.63\!\pm\!0.79$ & $38.62\!\pm\!0.65$ & $32.31\!\pm\!0.55$ \\
FedKD$^*$ & $40.10\!\pm\!0.19$ & $70.27\!\pm\!1.19$ & $38.54\!\pm\!0.54$ & $32.23\!\pm\!0.54$ \\
Naive & $39.35\!\pm\!0.38$ & $70.14\!\pm\!0.73$ & $41.07\!\pm\!0.64$ & $34.48\!\pm\!0.22$ \\
\bottomrule
\end{tabular*}
\endgroup
\end{table}

\subsection{Deployment Gains at Matched Inference Capacity}
\begin{table}[t]
\centering\footnotesize\setlength{\tabcolsep}{3pt}
\caption{Joint outperforms parameter-matched independent ensembles (\%; mean $\pm$ SD, five seeds). UCF uses label-trained Private branches and pooled accuracy; Flickr uses Q-trained Private branches and I2T/T2I R@1. Full configurations, costs, and DEV weights: \cref{tab:ensemble-controls-full}.}
\label{tab:capacity-main}\label{tab:ensemble-controls-main}
\begin{tabular*}{\columnwidth}{@{\extracolsep{\fill}}lccc@{}}
\toprule
Predictor & UCF Acc. & I2T R@1 & T2I R@1 \\
\midrule
Joint & \best{$82.83\!\pm\!0.93$} & \best{$42.37\!\pm\!0.51$} & \best{$35.46\!\pm\!0.48$} \\
Matched ensemble & $71.77\!\pm\!0.17$ & $36.56\!\pm\!0.21$ & $31.54\!\pm\!0.05$ \\
\bottomrule
\end{tabular*}
\end{table}
At matched inference capacity, Joint outperforms parameter-matched independent ensembles by $11.06\pp$ in UCF pooled accuracy and $4.87\pp$ in mean bidirectional Flickr30k R@1 (\cref{tab:capacity-main}). UCF Joint and its matched ensemble each use 0.60M parameters and 0.012 ms/sample at displayed precision. Flickr Joint and the independent Q-trained Private ensemble each use 36.15M parameters, with 7.27 versus 7.29 GFLOPs and 0.260 versus 0.292 ms/sample. Q provides a stronger deployment partner at the same parameter budget, with positive UCF gains in every seed. Flickr uses five cyclic combinations of training replicas (\cref{tab:ensemble-controls-paired}).

The controls use the same local data and DEV-selected mixtures of matched branches, two full-sized Private models, or checkpoints (\cref{tab:ensemble-controls-full}). Both Flickr ensemble branches retain Q-assisted training, isolating the value of keeping Q at deployment.

\subsection{Cross-Task Sharing and Bidirectional Teaching}
\begin{figure}[t]
\centering
\includegraphics[width=0.88\linewidth]{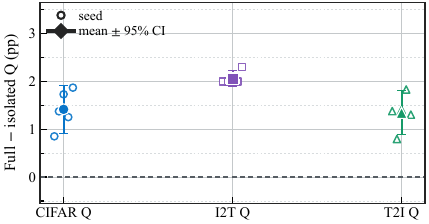}
\caption{Overall contribution of cross-task sharing: Full Q minus task-isolated Q, which retains within-task federation (pp). Small symbols: paired seed differences; large symbols: means; bars: 95\% paired $t$ confidence intervals (CIs). CIFAR: accuracy; Flickr I2T/T2I: image-to-text/text-to-image R@1.}
\label{fig:sharing}
\end{figure}

Cross-task sharing strengthens Q beyond within-task federation. Full exceeds task-isolated Q by $1.42\pm0.40\pp$ on CIFAR and $2.06\pm0.13\pp$/$1.34\pm0.37\pp$ on Flickr I2T/T2I, positive in all five pairs (\cref{tab:mechanism-contrasts,fig:sharing}). The control retains within-task training and aggregation, so these gains measure the overall benefit of exchanging knowledge across tasks while preserving their prediction objectives. \Cref{tab:mechanism-absolute} gives complete endpoints.

Bidirectional teaching strengthens both sides of the exchange. \emph{Writing local knowledge into Q} improves Q over Label Q by $1.41\pp$ on CIFAR and $2.64\pp$/$2.07\pp$ on Flickr I2T/T2I. Private also improves over Label Q by $1.31\pm0.57\pp$ and $1.11\pm0.14\pp$/$1.19\pm0.48\pp$, respectively. \emph{Returning aggregated knowledge to Private} improves Private over No return by $6.58\pp$ on CIFAR and $16.87\pp$/$15.29\pp$ on Flickr. Local predictions enrich Q, and the returned Q makes that federated knowledge useful to private models.

\begin{table}[t]
\centering\footnotesize\setlength{\tabcolsep}{3pt}
\caption{Complete KD$\times$fusion factorial (\%; mean $\pm$ SD, five seeds). Neither/KD/Fusion/Full enable neither term, KD alone, fusion alone, or both. All retain Private-to-Q teaching. CIFAR/AG: accuracy; I2T/T2I: Flickr R@1. Red/underline: best/second-best within each row. Effects: \cref{tab:kd-fusion-effects}.}
\label{tab:kd-fusion-absolute}
\setlength{\tabcolsep}{1.5pt}
\begin{tabular*}{\linewidth}{@{\extracolsep{\fill}}lcccc@{}}
\toprule
Metric & Neither & KD & Fusion & Full \\
\midrule
\multicolumn{5}{@{}l}{\emph{Private}} \\
CIFAR & $38.55\!\pm\!0.13$ & \second{$41.86\!\pm\!0.14$} & $38.56\!\pm\!0.05$ & \best{$45.13\!\pm\!0.39$} \\
AG News & $46.75\!\pm\!0.15$ & \second{$48.79\!\pm\!0.41$} & $47.13\!\pm\!0.29$ & \best{$49.73\!\pm\!0.48$} \\
I2T R@1 & $18.22\!\pm\!0.74$ & \best{$35.10\!\pm\!0.40$} & $17.59\!\pm\!0.53$ & \second{$35.09\!\pm\!0.26$} \\
T2I R@1 & $15.08\!\pm\!0.26$ & \second{$30.24\!\pm\!0.34$} & $14.51\!\pm\!0.36$ & \best{$30.38\!\pm\!0.45$} \\
\addlinespace[2pt]
\multicolumn{5}{@{}l}{\emph{Joint}} \\
CIFAR & $45.62\!\pm\!0.95$ & \second{$49.95\!\pm\!0.97$} & $46.23\!\pm\!0.72$ & \best{$53.59\!\pm\!1.28$} \\
AG News & $70.19\!\pm\!0.82$ & \second{$74.14\!\pm\!0.66$} & $71.07\!\pm\!0.73$ & \best{$78.50\!\pm\!0.42$} \\
I2T R@1 & $37.41\!\pm\!0.58$ & \second{$40.65\!\pm\!0.60$} & $37.13\!\pm\!0.75$ & \best{$42.37\!\pm\!0.51$} \\
T2I R@1 & $31.39\!\pm\!0.28$ & \second{$34.67\!\pm\!0.20$} & $31.24\!\pm\!0.21$ & \best{$35.46\!\pm\!0.48$} \\
\bottomrule
\end{tabular*}

\end{table}

\subsection{Interactions That Strengthen Joint Prediction}
The complete KD$\times$fusion factorial tests how learning from Q and training the combined prediction work together (\cref{tab:kd-fusion-absolute,tab:kd-fusion-effects}). KD has positive average effects on every Private and Joint metric; fusion has positive average effects on all Joint metrics. Their Joint interactions are positive across all four metrics. On AG, fusion adds $0.88\pp$ without KD and $4.36\pp$ with KD, producing a $3.48\pp$ interaction. Learning from Q therefore increases the deployment benefit of training Private and Q to predict together.

\begin{figure}[t]
\centering
\includegraphics[width=\linewidth]{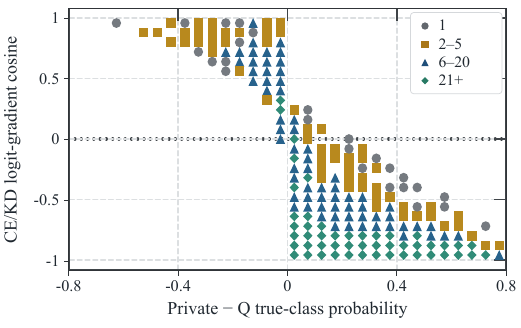}
\caption{Recipient confidence shapes supervision agreement. CREMA-D TRAIN, 6,096 correct-teacher observations: horizontal is Private minus Q true-class probability; vertical is the cosine between CE and KD gradients at private logits. Symbols count observations per occupied bin (1, 2--5, 6--20, 21+). Negative cosine denotes opposing output-loss directions. Full development protocol: \cref{sec:development-diagnostics}.}
\label{fig:recipient-supervision}
\end{figure}

\section{Analysis and Discussion}
\paragraph{Supervision at the recipient.}
\Cref{fig:recipient-supervision} shows why teacher correctness and recipient compatibility are distinct. When Private assigns greater true-class probability than Q, label and distillation gradients can oppose each other. This motivates examining transferred knowledge through recipient learning and prediction; \cref{sec:development-diagnostics} gives the development diagnostics.

\paragraph{Evaluating the prediction partner.}
A separate two-stage CREMA-D development study selects Q, then recipient training and fusion (\cref{fig:diag-d03}). Conditional Q attains the highest mean Joint accuracy, $76.60\pm0.42\%$, versus its Private and Q endpoints at $76.28\%$ and $69.44\%$. The audit uses 430 examples held from previously observed DEV; \cref{sec:development-diagnostics} details selection. This endpoint view evaluates how Q complements local prediction.

\paragraph{Deployment across private architectures.}
Joint improves mean macro F1 and UAR across all three UCF backbone groups (\cref{tab:followup-backbones}), reaching $10.92\pp$/$10.68\pp$ gains for Transformers. Multilayer perceptron (MLP) and convolutional neural network (CNN) groups gain $15.02\pp$/$18.57\pp$ accuracy, positive in every seed, while retaining their private architectures.

\section{Conclusion}
FedSocket connects heterogeneous private models through Q by jointly designing recipient-computable inputs, task-owned outputs, and compatible aggregation. Teacher and task-isolated controls establish the benefits of complete-modality supervision and cross-task exchange. Across six datasets, Joint prediction improves missing-modality recipient accuracy over Local by $14.44\pp$ on MELD and $15.51\pp$ on UCF-51. FedSocket leads FML/FedKD adaptations and Naive Shared on all four primary classification and retrieval metrics. At matched inference capacity, Joint outperforms independent ensembles by $11.06\pp$ in UCF-51 pooled accuracy and $4.87\pp$ in mean bidirectional Flickr30k R@1. Recipient-executable exchange turns federated knowledge into stronger predictions.

\par
\clearpage
\nobalance

{
    \footnotesize
    \setlength{\bibsep}{0pt plus 0.2pt minus 0.2pt}
    \bibliographystyle{ieeenat_fullname}
    \bibliography{main}
}
\clearpage
\appendix
\setcounter{topnumber}{3}
\setcounter{bottomnumber}{2}
\setcounter{totalnumber}{5}
\setcounter{dbltopnumber}{3}
\renewcommand{\topfraction}{0.95}
\renewcommand{\bottomfraction}{0.90}
\renewcommand{\dbltopfraction}{0.95}
\renewcommand{\textfraction}{0.05}
\clubpenalty=10000
\widowpenalty=10000
\displaywidowpenalty=10000
\setlength{\textfloatsep}{10pt plus 2pt minus 2pt}
\setlength{\dbltextfloatsep}{10pt plus 2pt minus 2pt}
\setlength{\floatsep}{8pt plus 2pt minus 2pt}
\setlength{\dblfloatsep}{8pt plus 2pt minus 2pt}
\setlength{\intextsep}{8pt plus 2pt minus 2pt}

\section{Complete Objectives and Implementation}
\label{sec:method-details}
\subsection{Recipient-Conditioned Task Projection}
Let $F_k$ return donor $k$'s class logits, $T>0$ be the distillation temperature, and $\softmax$ convert logits to a probability vector. The temperature-scaled teacher distribution is
\begin{equation}
t_k(x_S,x_U)=\softmax\!\left(F_k(x_S,x_U)/T\right).
\end{equation}
For a fixed donor population $D$, the target predicted from $z=h_S(x_S)$ is
\begin{equation}
Q_S^*(z)=\E_D\!\left[t_k(x_S,x_U)\mid h_S(x_S)=z\right].
\label{eq:conditional-target-detail}
\end{equation}
Equation~\eqref{eq:conditional-target} defines the optimal prediction expressible through the recipient-observed interface for a fixed within-task teacher under unrestricted forward-KL risk~\cite{banerjee2005conditional}. The conditional target averages teacher predictions over examples sharing the same interface value. FedSocket approximates this projection using teacher predictions, labels, and reliability weights (\cref{sec:projection-proof}).

\subsection{Interfaces and Private Models}
Q reads fixed local features computed independently of private encoders. Same-task experiments use precomputed text or visual features. Task-heterogeneous experiments use frozen ResNet-18 image features and GloVe-mean text features with a fixed orthogonal projection, both in $\mathbb R^{256}$. \Cref{sec:implementation-details} lists feature dimensions and donor backbones.

Private models remain local: same-task recipients use multilayer perceptron (MLP), temporal convolutional neural network (CNN), or Transformer variants; CIFAR-100 and AG News use ResNet-18/PIE and GloVe--BiGRU/PIE encoders. Flickr30k uses independent modality encoders and normalized 256-dimensional embeddings; its private score is scaled cosine similarity, with all captions of an image treated as positives.

\subsection{FedSocket Architecture}
For task $t$ and modality $m$, let $u=h_m(x^m)$ be the fixed input feature. Q applies adapter $A_{t,m}$ and adds learned task embedding $e_t$ and modality embedding $e_m$ to form intermediate feature $b$:
\begin{equation}
b=A_{t,m}(u)+e_t+e_m.
\end{equation}
The hidden representation combines a task-private expert $E_t$ with a gated low-rank shared residual $R$:
\begin{equation}
v_{t,m}(u)=E_t(b)+\sigma(g_t)R(b).
\label{eq:qhidden-detail}
\end{equation}
Here $v_{t,m}$ is the hidden feature, $g_t$ is a learned task-specific scalar, and sigmoid $\sigma$ maps it to the residual weight. The hidden width is 256 and the shared residual rank is 48. Classification tasks use separate linear heads; Flickr30k uses a shared 256-dimensional normalized retrieval head. The server aggregates task-specific components within each task and shares $R$ and modality embeddings across compatible tasks.

\subsection{Stage I: Private Warm-Up and Same-Task Cross-Fitting}
Private models are first trained with their native supervised objectives. In the cross-fitted same-task protocol, donor training groups are split into two folds by dialogue or video group. A full-modality teacher trained without the held group produces out-of-fold (OOF) logits $\ell_i^{\mathrm{OOF}}$ for example $i$. The teacher contains an observed-modality public prior plus a bounded private correction:
\begin{equation}
\ell_i^{\mathrm{OOF}}=Q_{\mathrm{prior}}(v_i)
+1.5\tanh\!\left(\frac{r_k(v_i,a_i)-\overline r_k}{3}\right).
\end{equation}
Here $v_i$ is the observed-modality feature (text or video), $a_i$ is the audio feature, $Q_{\mathrm{prior}}$ supplies prior logits, and $r_k$ is client $k$'s learned logit correction. Its centering statistic $\overline r_k$ is computed outside the held group; $\tanh$ is applied elementwise, and 1.5 and 3 set the correction bound and scale. Cached logits remain local. Teacher training excludes the held group from every label-trained prior, residual component, and centering statistic used to produce its targets.

\subsection{Stage II: Q as Teacher of the Private Model}
At round $r$, the server broadcasts $Q(\cdot;\phi^r)$. Each client freezes teacher copy $Q^T$ and initializes student copy $Q^S$ from it. Let $p_k$ and $q_k^{\mathrm{teach}}$ denote Private and teacher-Q classification logits. This stage updates Private. For example $i$, teacher confidence $\gamma_i$ and reliability weight $w_i$ use Shannon entropy $H$ over $C$ classes:
\begin{align}
\gamma_i(q) &= 1-\frac{H(\softmax(q_i))}{\log C},\\
w_i(q,y) &= \gamma_i(q)\left(0.25+0.75\mathbf{1}[\arg\max q_i=y_i]\right).
\end{align}
For probability vector $\pi$, $H(\pi)=-\sum_c\pi_c\log\pi_c$, where $c$ indexes classes. The indicator $\mathbf 1[\cdot]$ equals one when its condition holds, and $\arg\max q_i$ is the predicted class. With $w_i=w_i(q,y)$, the reliability-weighted distillation loss is
\begin{equation}
\begin{aligned}
\mathcal{L}_{\mathrm{RKD}}(p,q;y)
&=\frac{1}{\sum_iw_i+\epsilon}\sum_iw_iT^2\\
&\quad\cdot\KL\!\left(\softmax(q_i/T)\|\softmax(p_i/T)\right).
\end{aligned}
\end{equation}
Here $p_i,q_i$ are student and teacher logit vectors for example $i$, sums run over the minibatch, and $\epsilon>0$ stabilizes the denominator. For probability vectors $a,b$, $\KL(a\|b)=\sum_c a_c\log(a_c/b_c)$. With label-smoothed cross-entropy $\CE_{\mathrm{ls}}$, the private classification objective (superscript $\mathrm{cls}$) is
\begin{equation}
\begin{aligned}
\mathcal{L}^{\mathrm{cls}}_{\mathrm{priv}}
&=\CE_{\mathrm{ls}}(p_k,y)\\
&\quad+\lambda_{\mathrm{KD}}\mathcal{L}_{\mathrm{RKD}}
   (p_k,\sg(q_k^{\mathrm{teach}});y)\\
&\quad+\lambda_{\mathrm{fus}}\mathcal L_{\mathrm{fus}}(p_k,q_k^{\mathrm{teach}},y).
\end{aligned}
\label{eq:private-class}
\end{equation}
Classification fusion operates in probability space. Let $\pi_k^P=\softmax(p_k)$ and $\pi_k^Q=\softmax(q_k^{\mathrm{teach}})$ be Private and frozen-Q class probabilities. Each client learns scalar $a_k$, initialized to $-0.5$, jointly with Private; its Q mixing weight $g_k^{\mathrm{train}}=\sigma(a_k)$ initially equals approximately $0.3775$. The mixture probability for class $c$ is $\pi_{k,c}^{\mathrm{fus}}$:
\begin{equation}
\begin{aligned}
\pi_{k,c}^{\mathrm{fus}}&=(1-g_k^{\mathrm{train}})\pi_{k,c}^P
 +g_k^{\mathrm{train}}\sg(\pi_{k,c}^Q),\\
\mathcal L_{\mathrm{fus}}&=-\frac{1}{B}\sum_{i=1}^B
\log\pi_{k,y_i}^{\mathrm{fus}}(x_i).
\end{aligned}
\label{eq:classification-fusion-training}
\end{equation}
Here $B$ is the batch size, $\pi_{k,y_i}^{\mathrm{fus}}(x_i)$ is the mixture probability assigned to the true label of example $i$, and $\lambda_{\mathrm{fus}}=0.25$. Fusion mixes softmax probabilities; KD uses the temperature-scaled KL defined above. The client mixing weight $g_k^{\mathrm{train}}$ operates on predictions; Q's task gate $\sigma(g_t)$ operates on hidden features.

For retrieval, $S^P$ and $S^Q$ are Private and Q image--text similarity matrices: rows index images and columns index captions in a minibatch. Q stacks unit-normalized image and text embeddings in $Z^{\mathrm{img}}$ and $Z^{\mathrm{txt}}$, giving $S^Q=15Z^{\mathrm{img}}(Z^{\mathrm{txt}})^\top$. The superscript $\top$ denotes transpose and 15 is the similarity scale. The retrieval objective (superscript $\mathrm{ret}$) is
\begin{equation}
\begin{aligned}
\mathcal{L}^{\mathrm{ret}}_{\mathrm{priv}}
&=\mathcal{L}_{\mathrm{MPNCE}}(S^P)\\
&\quad+\lambda_{\mathrm{KD}}\mathcal{L}_{\mathrm{symKL}}
   (S^P,\sg(S^Q))\\
&\quad+\lambda_{\mathrm{fus}}\mathcal{L}_{\mathrm{MPNCE}}
   ((1-g_{\mathrm{ret}}^{\mathrm{train}})S^P+g_{\mathrm{ret}}^{\mathrm{train}}S^Q).
\end{aligned}
\label{eq:private-ret}
\end{equation}
Here $\mathcal L_{\mathrm{MPNCE}}$ is multi-positive contrastive loss, treating all captions of the same image as positives; $\mathcal L_{\mathrm{symKL}}$ is the symmetric KL distillation loss over image-to-text and text-to-image score distributions. The scalar $g_{\mathrm{ret}}^{\mathrm{train}}\in[0,1]$ weights Q scores in training-time fusion. KD matches Private to the frozen Q distributions; fusion supervises their combined scores with the true matches. Both terms update Private. Joint denotes inference with both predictors; its mixing coefficients are selected separately on DEV.

\subsection{Stage III: Q as Student of the Private Model}
After Private updates, client $k$ trains its local student $Q^S(\cdot;\phi_k^r)$, where $\phi_k^r$ denotes its local Q parameters in round $r$, initialized from global $\phi^r$. Private teacher outputs are detached. Student logits are $q_k$; client-specific Private and student retrieval matrices are $S_k^P,S_k^Q$. For classification,
\begin{equation}
\mathcal{L}^{\mathrm{cls}}_{Q,k}
=\CE_{\mathrm{ls}}(q_k,y)
+\lambda_{\mathrm{rel}}\mathcal{L}_{\mathrm{RKD}}(q_k,\sg(p_k);y).
\label{eq:q-class}
\end{equation}
For retrieval,
\begin{equation}
\mathcal{L}^{\mathrm{ret}}_{Q,k}
=\mathcal{L}_{\mathrm{MPNCE}}(S_k^Q)
+\lambda_{\mathrm{rel}}\mathcal{L}_{\mathrm{symKL}}(S_k^Q,\sg(S_k^P)).
\label{eq:q-ret}
\end{equation}
In \cref{eq:private-class,eq:private-ret}, Q teaches Private; in \cref{eq:q-class,eq:q-ret}, Private teaches Q. The second argument of $\mathcal L_{\mathrm{RKD}}$ provides the teacher distribution and reliability weight, and $\sg$ stops its gradients. Symbols $p,q$ denote logits. Fusion training precedes DEV selection of the inference weights.

For the same-task cross-fitted protocol, local Q instead combines a label anchor and reliable out-of-fold teacher correction:
\begin{equation}
\begin{aligned}
\mathcal{L}^{\mathrm{same}}_{Q,k}
&=\CE(q_k,y)\\
&\quad+\lambda_{\mathrm{same}}\mathbf{1}[\arg\max \ell^{\mathrm{OOF}}=y]T^2\\
&\qquad\cdot\KL(t_T^{\mathrm{OOF}}\|\pi_{k,T}^{Q}).
\end{aligned}
\end{equation}
Here $\CE$ is ordinary cross-entropy, $\lambda_{\mathrm{same}}\geq0$ weights same-task teacher supervision, $t_T^{\mathrm{OOF}}=\softmax(\ell^{\mathrm{OOF}}/T)$ is the OOF teacher distribution, and $\pi_{k,T}^{Q}=\softmax(q_k/T)$ is the student's temperature-scaled distribution. The loss is averaged over local examples. Recipients train from observed inputs using local labels and a reliability-gated global-Q distillation term.

\subsection{Heterogeneous Server Aggregation}
Clients upload local Q parameters $\phi_k^r$, training-sample count $n_k$, and, for classification, class-count vector $c_k=(c_{k,c})_c$, whose entry $c_{k,c}$ counts training examples of class $c$. Raw samples, private features, private parameters, and sample-level logits remain local. Adapters $A_{t,m}$, experts $E_t$, task embeddings $e_t$, gates $g_t$, and heads are aggregated within their task. Tasks with modality $m$ share $e_m$; all tasks contribute to residual $R$. A classification-head row uses clients with positive counts for that class.

For residual parameters $\phi_R$, let $\Delta_t$ denote the task-level average update and initialize $\widetilde\Delta_t=\Delta_t$.  Conflicting task directions are projected in a PCGrad-style merge~\cite{yu2020pcgrad}:
\begin{equation}
\widetilde{\Delta}_t\leftarrow
\widetilde{\Delta}_t-
\frac{\min(0,\langle\widetilde{\Delta}_t,\Delta_s\rangle)}
{\|\Delta_s\|_2^2+\epsilon}\Delta_s,
\quad s\ne t,
\end{equation}
Here $s$ indexes another task in $\mathcal T$, $\langle\cdot,\cdot\rangle$ is the Euclidean inner product, $\|\cdot\|_2$ its norm, and $\epsilon>0$ stabilizes the projection. The server then uses $\phi_R^{r+1}=\phi_R^r+|\mathcal{T}|^{-1}\sum_t\widetilde{\Delta}_t$, averaging projected updates across tasks. Modality embeddings merge contributions from tasks using that modality. Image embeddings connect CIFAR-100/Flickr30k; text embeddings connect AG News/Flickr30k.

The same-task implementation uses sample-weighted Q averaging followed by a 0.5 server stabilizer. Task-update projection applies to the task-heterogeneous instance.

\subsection{Development Calibration and Inference}
We report three endpoints: \emph{Private} uses $f_k$, \emph{public Q} uses Q, and \emph{Joint} combines them. For non-IID classification, the local training histogram determines a class-specific Q weight and score
\begin{equation}
\begin{aligned}
\omega_{k,c}&=\frac{\kappa_k}{c_{k,c}+\kappa_k},\\
s_{k,c}^{\mathrm{joint}}&=(1-\omega_{k,c})\pi_{k,c}^{P}
+\omega_{k,c}\pi_{k,c}^{Q}.
\end{aligned}
\label{eq:inference-fusion}
\end{equation}
Here $\omega_{k,c}$ weights Q for class $c$, $\kappa_k>0$ controls its dependence on local count $c_{k,c}$, and $\pi_{k,c}^{P},\pi_{k,c}^{Q}$ are Private and aggregated-Q probabilities. An absent class has $\omega_{k,c}=1$. DEV selects $\kappa_k$ from $\{0.02,0.05,0.1,0.2,0.5,1,2,5,10^6\}$. The resulting Joint score $s_{k,c}^{\mathrm{joint}}$ is normalized across classes for probability evaluation, leaving its argmax unchanged. Inference uses these class-dependent weights in place of the learned training scalar. Flickr30k mixes $S^P$ and $S^Q$ into Joint scores $S^J$, with separate DEV-selected weights for each retrieval direction.

For the same-task UCF protocol, Joint probability vector $\pi^{J}$ instead uses scalar Q weight $\beta$: $\pi^{J}=(1-\beta)\pi^P+\beta\pi^Q$. Recipient DEV accuracy selects $\beta$ from $\{0,0.2,0.35,0.5,0.75,1\}$, using macro F1 to break ties, and the selected value is frozen for TEST. Each method and seed uses the same calibration grid. Dirichlet concentration $\alpha=0.5$ controls training-client label partitions.

\subsection{Training and Deployment Protocol}
Each round broadcasts Q, updates Private, trains a local Q copy, and aggregates Q blocks according to ownership. Private inference retains $f_k$; Q retains $h_m$ and Q; Joint retains both predictors and the fixed fusion rule. Clients are registered during training. Only Q parameters and permitted counts are communicated. \Cref{sec:implementation-details} gives payload accounting and implementation details.

\section{What Related Methods Exchange}
\Cref{tab:capability} compares the inputs, uploads, output spaces, and inference options of the closest methods.
\begin{table*}[t]
\caption{Inputs, communication, and inference in closely related methods. ``Sample-indexed upload'' means logits or representations associated with shared examples. ``Recipient branch'' means a returned shared model with inputs the recipient can compute and outputs valid for its task. Entries describe the published methods and evaluated inference options.}
\label{tab:capability}
\centering
\footnotesize
\setlength{\tabcolsep}{5.2pt}
\begin{tabular}{@{}lccccc@{}}
\toprule
Method & Common public examples & Sample-indexed upload & Output organization & Recipient branch & Evaluated inference \\
\midrule
FedMD~\cite{li2019fedmd} & Required & Logits & Common output space & No & Private \\
FML~\cite{shen2020fml} & Not required & None & Adapted client heads & Training meme & Private/global \\
FedKD~\cite{wu2022fedkd} & Not required & None & Common task outputs & Yes (same task) & Private mentor \\
MH-pFLID~\cite{xie2024mhpflid} & Not required & None & Receiver/transmitter modules & Yes & Private \\
CreamFL~\cite{yu2023creamfl} & Required & Representations & Global server space & No & Client/server \\
FedHTCM~\cite{yin2025fedhtcm} & Required & Representations & Multitask global space & No & Client/global \\
FedAFD~\cite{tan2026fedafd} & Required & Representations & Client/server task models & No & Client/server \\
\textbf{FedSocket} & \textbf{Not required} & \textbf{None} & \textbf{Task-owned heads} & \textbf{Yes} & \textbf{Private/Q/Joint} \\
\bottomrule
\end{tabular}
\end{table*}

\captionsetup{hypcap=false}
\section{Population Interpretation of the Conditional Interface}
\label{sec:projection-proof}
Let $X$ be the random complete-modality input, $Y$ its label, and $X_S$ the recipient-observed part. A fixed teacher returns probability vector $t(X)$, and $Z=h_S(X_S)$ is the recipient feature, with realization $z$. Expectations $\E$ in this projection are under donor population $D$. For a measurable probability predictor $q(Z)$ with finite expected forward-KL risk, define $\mathcal R(q)$ by
\begin{equation}
\mathcal R(q)=\E\big[\KL(t(X)\|q(Z))\big].
\end{equation}
Writing the conditional mean teacher vector as $\mu(Z)=\E[t(X)\mid Z]$ and conditioning on $Z$ gives
\begin{equation}
\mathcal R(q)=\mathcal R(\mu)+\E\big[\KL(\mu(Z)\|q(Z))\big].
\end{equation}
To verify this identity, expand both KL terms and use
$\E[t_c(X)\log q_c(Z)]=\E[\mu_c(Z)\log q_c(Z)]$ for every class $c$.
Nonnegativity of KL gives $q^*(Z)=\mu(Z)$ almost surely. This conditional Bregman result~\cite{banerjee2005conditional} defines the optimal population target for the recipient-observed interface. The projection identifies the teacher information expressible through the recipient interface.

Label supervision and reliability weighting extend the projection to a label-and-teacher mixture. Let $w(X,Y)\geq0$ be teacher reliability, and let $a,b\geq0$ weight labels and teacher supervision in this population calculation. Here $\CE(v,q)=-\sum_c v_c\log q_c$ takes a target probability vector first. Consider the simplified common-temperature objective $\mathcal L(q)$:
\begin{equation}
\begin{aligned}
\mathcal L(q)=\E\big[&a\,\CE(e_Y,q(Z))\\
&+b\,w(X,Y)\CE(t(X),q(Z))\big],
\end{aligned}
\end{equation}
Here $e_Y$ is the one-hot vector for label $Y$, and $\Pr(Y=\cdot\mid Z=z)$ denotes the vector of conditional class probabilities. Provided the denominator is positive, the unrestricted minimizer $q^*$ is
\begin{equation}
q^*(z)=\frac{\begin{gathered}a\,\Pr(Y=\cdot\mid Z=z)\\+b\,\E[w(X,Y)t(X)\mid Z=z]\end{gathered}}
{a+b\,\E[w(X,Y)\mid Z=z]}.
\end{equation}
Minimizing conditional cross-entropy gives a mixture of labels and reliability-weighted teacher predictions. The implemented approximation uses temperature scaling, minibatch normalization, and finite-model optimization.

\paragraph{Recipient-label risk.}
For a recipient population, denoted by subscript $\mathrm{rec}$, let $\eta_{\mathrm{rec}}(z)=\Pr_{\mathrm{rec}}(Y=\cdot\mid Z=z)$ be its label-conditional vector in the same label space. Write $\mu_D(z)=\E_D[t(X)\mid Z=z]$ for the donor projection. Assume $P_{\mathrm{rec}}^Z\ll P_D^Z$: the recipient distribution of $Z$ assigns zero mass to every set with zero donor mass. Also assume finite cross-entropies. Define recipient label risk $\mathcal C_{\mathrm{rec}}(q)=\E_{\mathrm{rec}}[-\log q_Y(Z)]$, where $q_Y$ selects the true-label probability. Conditioning on $Z$ gives
\begin{equation}
\begin{aligned}
&\mathcal C_{\mathrm{rec}}(\mu_D)-\mathcal C_{\mathrm{rec}}(\eta_{\mathrm{rec}})\\
&\quad=\E_{\mathrm{rec}}\big[\KL(\eta_{\mathrm{rec}}(Z)\|\mu_D(Z))\big].
\end{aligned}
\label{eq:recipient-label-risk}
\end{equation}
A sufficient condition for zero mismatch is $t(X)=\Pr_D(Y=\cdot\mid X)$ and $\Pr_D(Y=\cdot\mid Z)=\eta_{\mathrm{rec}}(Z)$. The tower property then gives $\mu_D=\eta_{\mathrm{rec}}$. Equation~\eqref{eq:recipient-label-risk} expresses recipient excess cross-entropy as the divergence between its label conditional and the donor projection, identifying the population quantities relevant to transfer.

\paragraph{How label anchoring limits population mismatch.}
For label anchoring, define mean reliability $W(z)=\mathbb E_D[w\mid Z=z]$, weighted teacher vector $\mu_w(z)=\mathbb E_D[wt\mid Z=z]/W(z)$ for $W(z)>0$, and teacher mixture weight $\lambda(z)=bW(z)/(a+bW(z))$. Assume the donor label conditional equals $\eta_{\mathrm{rec}}(z)$ on recipient support and $a>0$. The anchored population target, denoted $q_a$, is
\begin{equation}
q_a(z)=(1-\lambda(z))\eta_{\mathrm{rec}}(z)+\lambda(z)\mu_w(z).
\end{equation}
Suppressing the argument $z$, let $q_{a,c}$ and $\eta_{\mathrm{rec},c}$ denote class-$c$ entries. Convexity of KL in its second argument and $q_{a,c}\geq(1-\lambda)\eta_{\mathrm{rec},c}$ yield
\begin{equation}
\begin{aligned}
\mathcal C_{\mathrm{rec}}(q_a)-\mathcal C_{\mathrm{rec}}(\eta_{\mathrm{rec}})
\leq\mathbb E_{\mathrm{rec}}\!\left[\min\left\{
\begin{gathered}
\lambda\,\mathrm{KL}(\eta_{\mathrm{rec}}\|\mu_w),\\
-\log(1-\lambda)
\end{gathered}\right\}\right].
\end{aligned}
\label{eq:anchor-bound}
\end{equation}
For $W=0$, set $\lambda=0$ and $q_a=\eta_{\mathrm{rec}}$. Under label-conditional alignment, convexity gives the first bound and the coordinatewise lower bound on $q_a$ gives the second. Label anchoring controls the teacher contribution through $\lambda$. Under label shift, the anchor uses donor label conditional $\eta_D(z)=\Pr_D(Y=\cdot\mid Z=z)$; recipient risk then depends on its mismatch with $\eta_{\mathrm{rec}}$. The bound quantifies the role of label supervision in controlling the transferred teacher contribution.

\renewcommand{\arraystretch}{1.0}
\begin{table*}[tbp]
\centering\footnotesize\setlength{\tabcolsep}{3pt}
\caption{Supplementary endpoint metrics (\%; mean $\pm$ sample SD, five seeds). (a) Same-task recipients: accuracy, macro F1, and UAR. (b) Flickr five-fold retrieval at ranks 1/5/10; $^{\dagger}$Mean R is a point estimate. (c) Paired recipient accuracy gains in pp: $J-L=(P-L)+(J-P)$, with $L$, $P$, and $J$ denoting Local, Private, and Joint. Red/underline: best/second-best means within each endpoint.}
\label{tab:recipient-endpoints}\label{tab:retrieval-main}\label{tab:gain-accounting}
\textit{(a) Missing-modality recipient endpoints}\par\smallskip

\begin{tabular}{@{}llcccccc@{}}
\toprule
& & \multicolumn{3}{c}{Private} & \multicolumn{3}{c}{Joint} \\
\cmidrule(lr){3-5}\cmidrule(l){6-8}
Dataset & Method & Acc. & Macro F1 & UAR & Acc. & Macro F1 & UAR \\
\midrule
MELD & Local & \second{$53.81\pm6.54$} & $27.78\pm6.47$ & $28.51\pm5.46$ & $53.81\pm6.54$ & $27.78\pm6.47$ & $28.51\pm5.46$ \\
 & Label Q & $53.29\pm6.22$ & \second{$28.02\pm6.58$} & \second{$28.83\pm5.72$} & $66.97\pm2.21$ & $44.68\pm5.96$ & $43.74\pm4.83$ \\
 & Single-modal Q & $53.29\pm6.22$ & \second{$28.02\pm6.58$} & \second{$28.83\pm5.72$} & \second{$67.23\pm2.01$} & \second{$45.79\pm4.79$} & \second{$44.85\pm4.13$} \\
 & FedSocket & \best{$54.21\pm6.36$} & \best{$28.89\pm7.22$} & \best{$29.50\pm6.12$} & \best{$68.24\pm1.99$} & \best{$46.92\pm4.61$} & \best{$45.82\pm3.94$} \\
\addlinespace[3pt]
CREMA-D & Local & \second{$74.65\pm0.50$} & \second{$71.14\pm0.60$} & \second{$71.36\pm0.52$} & \second{$74.65\pm0.50$} & \second{$71.14\pm0.60$} & \second{$71.36\pm0.52$} \\
 & Label Q & $74.48\pm0.64$ & $70.28\pm0.72$ & $70.39\pm0.82$ & $74.48\pm0.64$ & $70.28\pm0.72$ & $70.39\pm0.82$ \\
 & Single-modal Q & $74.12\pm0.64$ & $69.97\pm0.99$ & $69.93\pm0.86$ & $74.12\pm0.64$ & $69.97\pm0.99$ & $69.93\pm0.86$ \\
 & FedSocket & \best{$75.05\pm0.95$} & \best{$71.54\pm1.27$} & \best{$71.76\pm1.32$} & \best{$75.24\pm0.46$} & \best{$71.44\pm0.44$} & \best{$71.60\pm0.41$} \\
\addlinespace[3pt]
UCF-51 & Local & $71.32\pm0.21$ & $66.41\pm0.42$ & $68.30\pm0.33$ & $71.32\pm0.21$ & $66.41\pm0.42$ & $68.30\pm0.33$ \\
 & Label Q & \second{$71.87\pm0.55$} & \second{$68.45\pm0.91$} & \second{$70.15\pm0.95$} & $84.62\pm1.12$ & $82.02\pm1.16$ & $85.30\pm0.79$ \\
 & Single-modal Q & $71.76\pm0.87$ & $67.92\pm1.24$ & $69.64\pm1.30$ & \second{$85.32\pm0.35$} & \second{$83.34\pm0.35$} & \second{$86.29\pm0.30$} \\
 & FedSocket & \best{$72.27\pm0.48$} & \best{$68.85\pm0.86$} & \best{$70.55\pm0.82$} & \best{$86.83\pm1.31$} & \best{$84.98\pm1.25$} & \best{$87.42\pm0.53$} \\
\bottomrule
\end{tabular}

\par\medskip
\begin{minipage}[t]{0.55\textwidth}
\vspace{0pt}\centering\textit{(b) Complete Flickr retrieval}\par\smallskip
\setlength{\tabcolsep}{3pt}
\begin{tabular}{@{}lccc@{}}
\toprule
Metric & Private & Public Q & Joint \\
\midrule
I2T R@1 & $35.09\pm0.26$ & \second{$37.86\pm0.82$} & \best{$42.37\pm0.51$} \\
I2T R@5 & $64.58\pm0.45$ & \second{$65.30\pm0.49$} & \best{$70.21\pm0.62$} \\
I2T R@10 & \second{$75.84\pm0.46$} & $75.76\pm0.43$ & \best{$80.26\pm0.46$} \\
T2I R@1 & $30.38\pm0.45$ & \second{$31.13\pm0.73$} & \best{$35.46\pm0.48$} \\
T2I R@5 & \second{$61.82\pm0.29$} & $61.47\pm0.56$ & \best{$66.93\pm0.32$} \\
T2I R@10 & \second{$74.65\pm0.22$} & $73.58\pm0.49$ & \best{$78.34\pm0.28$} \\
Mean R$^{\dagger}$ & $57.06$ & \second{$57.52$} & \best{$62.26$} \\
\bottomrule
\end{tabular}

\end{minipage}\hfill
\begin{minipage}[t]{0.42\textwidth}
\vspace{0pt}\centering\textit{(c) Private retention and Q deployment}\par\smallskip
\setlength{\tabcolsep}{2pt}
\begin{tabular}{@{}lccc@{}}
\toprule
Dataset & $P-L$ & $J-P$ & $J-L$ \\
\midrule
MELD & $0.40\pm1.26$ & $14.04\pm6.60$ & $14.44\pm6.59$ \\
CREMA-D & $0.40\pm1.17$ & $0.19\pm0.68$ & $0.59\pm0.76$ \\
UCF-51 & $0.95\pm0.46$ & $14.56\pm1.27$ & $15.51\pm1.35$ \\
\bottomrule
\end{tabular}

\end{minipage}
\end{table*}

\section{Experimental Protocol Details}
\label{sec:protocol-summary}\label{tab:protocol}
\Cref{tab:protocol-main} gives the six datasets and available inputs. Non-IID means client data are not independently and identically distributed across clients. MELD uses a selected four-class cohort and group cross-fitting; CREMA-D uses six emotions and same-actor sentence-group splits. UCF-51 has 20 clients, including six video--audio donors, with Dirichlet concentration $\alpha=0.5$. CIFAR/AG/Flickr have 3/3/4 non-IID clients. Heterogeneous classification averages fixed client endpoints; same-task All accuracy pools test samples; each Flickr seed aggregates five 200-image retrieval folds.

\paragraph{Replication and paired summaries.}
\label{sec:paired-summary}
Same-task seeds are $\{1,7,11,21,42\}$; task-heterogeneous and mechanism seeds are $\{101,202,303,404,505\}$. Training seeds are the independent replication units on fixed primary client partitions; client endpoints and retrieval folds are aggregated within each seed. Ablations use the primary Full configuration and its paired controls. Each paired difference uses corresponding Full/control seed records. For $n$ paired seeds, we enumerate all $2^n$ sign flips for the exact two-sided test; with $n=5$, its minimum $p$-value is $2/32=0.0625$. For seed $s$, let $d_s$ be the paired difference, $\bar d$ its sample mean, and $s_d$ its sample SD. Paired $t$ intervals use $\bar d\pm t_{0.975,n-1}s_d/\sqrt{n}$, where $t_{0.975,n-1}$ is the 97.5th percentile of Student's $t$ distribution with $n-1$ degrees of freedom. The $p$-value is the fraction of sign-flipped mean differences at least as extreme in magnitude as observed. Intervals are unadjusted per-contrast summaries without multiplicity correction. Positive-seed counts describe the direction of the measured differences. Flickr ensemble controls reuse replicas in cyclic combinations and are descriptive comparisons, with pairing specified in \cref{tab:ensemble-controls-full}.

\paragraph{Evaluation populations and aggregation.}
Same-task all-client accuracy, $A_s^{\mathrm{All}}$ in seed $s$, pools test samples. For client $k$ with $n_{k,s}$ test samples and accuracy $a_{k,s}$ in seed $s$, the endpoint is
\begin{equation}
 A_s^{\mathrm{All}}=\frac{\sum_{k\in\mathcal K}n_{k,s}a_{k,s}}
 {\sum_{k\in\mathcal K}n_{k,s}}.
\label{eq:pooled-accuracy}
\end{equation}
Mean and sample SD are computed across the five $A_s^{\mathrm{All}}$ values. Heterogeneous classification averages fixed client endpoints within each seed; Flickr aggregates five 200-image retrieval folds. Each method contrast uses the same metric and population.

\paragraph{Single-modal teacher control.}
Single-modal Q isolates donor audio information in Q supervision. MELD uses the donor's text encoder/head; CREMA-D and UCF use the video encoder/visual head. Conditional Q receives targets from the full text--audio or video--audio teacher. Donor Private remains multimodal in both controls, Q reads the same recipient-observed inputs, and labels supervise Q training.

\section{Supplementary Endpoint Metrics}
\label{sec:recipient-details}
\Cref{tab:recipient-endpoints} groups recipient class-balanced metrics, retrieval ranks, and the training/deployment gain decomposition. Local's Private and Joint columns denote its single predictor. Mean R is the point estimate obtained by averaging the six displayed retrieval recall means.

All-client accuracy pools donor and recipient test samples (\cref{eq:pooled-accuracy}). Donor/recipient counts are 1,807/155 for MELD, 349/837 for CREMA-D, and 554/1,390 for UCF. Cohort gaps summarize these distinct test populations; method gains compare the same recipient population within each dataset.

\section{Published Reference Context}
\Cref{tab:reference-heterogeneous} collects published results under their cited benchmark conditions and FedSocket's deployment endpoints. Protocol-matched shared-model comparisons appear in \cref{tab:neighbor-main,tab:neighbor-endpoints}.

\begin{table}[htbp]
\centering
\caption{Published reference results and FedSocket deployment endpoints. Classification and retrieval references follow the cited heterogeneous benchmark; same-task references use the indicated modality/metric settings. $^{\dagger}$AMID is a centralized UCF-51 reference.}
\label{tab:main_results}\label{tab:reference-heterogeneous}

\footnotesize
\setlength{\tabcolsep}{2pt}
\renewcommand{\arraystretch}{0.90}

\textit{(a) Classification and retrieval}\par
\vspace{1mm}

\begin{adjustbox}{max width=\linewidth}
\begin{tabular}{@{}lcccc@{}}
\toprule
Method & CIFAR & AG & I2T & T2I \\
\midrule
LOCAL~\cite{tan2026fedafd} & 28.07 & 48.35 & 22.33 & 18.44 \\
FedMD~\cite{li2019fedmd} & 22.54 & 48.18 & 19.13 & 15.63 \\
FedGEMS~\cite{cheng2021fedgems} & 22.84 & 48.30 & 18.93 & 16.05 \\
FedET~\cite{cho2022fedet} & 31.86 & 49.38 & 22.63 & 18.22 \\
CreamFL~\cite{yu2023creamfl} & 22.14 & 42.16 & 18.38 & 15.49 \\
FedMKD~\cite{li2024fedmkd} & 24.99 & 47.99 & 22.33 & 18.37 \\
FedDFA~\cite{wang2025feddfa} & 23.09 & 43.79 & 19.68 & 17.13 \\
FedAFD~\cite{tan2026fedafd} & 33.18 & 51.98 & 32.48 & 25.68 \\
\midrule
FedSocket (Private) & 45.13 & 49.73 & 35.09 & 30.38 \\
FedSocket (Joint) & 53.59 & 78.50 & 42.37 & 35.46 \\
\bottomrule
\end{tabular}
\end{adjustbox}

\vspace{2mm}

\textit{(b) Same-task all-client accuracy}\par
\vspace{1mm}

\begin{adjustbox}{max width=\linewidth}
\begin{tabular}{@{}lccc@{}}
\toprule
Dataset & Published ref. & Private & Joint \\
\midrule
MELD & MFedMC~\cite{yuan2024mfedmc} 53.31
& $61.11\pm1.13$ & $62.16\pm0.77$ \\
CREMA-D & BMSFed~\cite{fan2024bmsfed} 59.8
& $74.87\pm0.62$ & $75.45\pm0.34$ \\
UCF-51$^{\dagger}$ & AMID~\cite{chen2023amid} 73.8
& $72.50\pm0.29$ & $82.83\pm0.93$ \\
\bottomrule
\end{tabular}
\end{adjustbox}

\end{table}

\section{Paired Contributions and Training Effects}
\label{sec:paired-effects}
The paired summaries connect cross-task sharing, bidirectional teaching, and Joint deployment. The KD$\times$fusion factorial separates effects on Private and Joint; complete endpoints and factorial cells are reported in \cref{tab:mechanism-absolute,tab:kd-fusion-absolute}.

\begin{table}[htbp]
\centering\footnotesize\setlength{\tabcolsep}{1pt}
\caption{Paired contributions (pp; mean $\pm$ SD, five seeds). P: Private; Iso.: task-isolated Q; Label: Label Q; no ret.: no Q-to-Private training. All contrasts are positive in 5/5 seeds. CIFAR/AG: accuracy; I2T/T2I: Flickr R@1. Dashes: unreported.}
\label{tab:mechanism-contrasts}\label{tab:v4-paired}

\begin{tabular*}{\columnwidth}{@{\extracolsep{\fill}}lcccc@{}}
\toprule
Contrast & CIFAR & AG & I2T & T2I \\
\midrule
Q $-$ Iso. & $+1.42\!\pm\!0.40$ & -- & $+2.06\!\pm\!0.13$ & $+1.34\!\pm\!0.37$ \\
Q $-$ Label & $+1.41\!\pm\!0.39$ & -- & $+2.64\!\pm\!0.69$ & $+2.07\!\pm\!0.57$ \\
P $-$ no ret. & $+6.58\!\pm\!0.42$ & -- & $+16.87\!\pm\!0.63$ & $+15.29\!\pm\!0.49$ \\
\midrule
Joint $-$ P & $+8.46\!\pm\!1.12$ & $+28.77\!\pm\!0.17$ & $+7.28\!\pm\!0.54$ & $+5.09\!\pm\!0.47$ \\
Joint $-$ Q & $+3.27\!\pm\!1.68$ & $+2.65\!\pm\!0.90$ & $+4.50\!\pm\!0.41$ & $+4.33\!\pm\!0.82$ \\
\bottomrule
\end{tabular*}
\end{table}

\begin{table}[htbp]
\centering\footnotesize\setlength{\tabcolsep}{1pt}
\caption{Main effects and interactions of KD and fusion training (pp; paired mean $\pm$ SD, five seeds). $Y_{ab}$ is performance with the first/second factor switched off (0) or on (1) by $a/b$; interaction is $Y_{11}-Y_{10}-Y_{01}+Y_{00}$. Complete cells: \cref{tab:kd-fusion-absolute}.}
\label{tab:kd-fusion-effects}

\begin{tabular}{@{}lcccc@{}}
\toprule
Effect & CIFAR & AG & I2T & T2I \\
\midrule
\multicolumn{5}{@{}l}{\textit{Private: KD$\times$fusion}} \\
KD & $4.94\!\pm\!0.19$ & $2.32\!\pm\!0.26$ & $17.19\!\pm\!0.47$ & $15.51\!\pm\!0.42$ \\
Fusion & $1.64\!\pm\!0.25$ & $0.66\!\pm\!0.33$ & $-0.32\!\pm\!0.40$ & $-0.22\!\pm\!0.14$ \\
Interaction & $3.26\!\pm\!0.57$ & $0.56\!\pm\!0.88$ & $0.62\!\pm\!0.68$ & $0.71\!\pm\!0.74$ \\
\midrule
\multicolumn{5}{@{}l}{\textit{Joint: KD$\times$fusion}} \\
KD & $5.84\!\pm\!1.36$ & $5.69\!\pm\!0.43$ & $4.24\!\pm\!0.30$ & $3.76\!\pm\!0.43$ \\
Fusion & $2.13\!\pm\!0.72$ & $2.62\!\pm\!0.42$ & $0.72\!\pm\!0.18$ & $0.32\!\pm\!0.26$ \\
Interaction & $3.03\!\pm\!1.25$ & $3.48\!\pm\!0.59$ & $2.00\!\pm\!0.58$ & $0.94\!\pm\!0.27$ \\
\bottomrule
\end{tabular}
\end{table}

\section{Absolute Cross-Task Sharing and Supervision Controls}
\label{sec:mechanism-details}
\Cref{tab:mechanism-absolute} reports Private, Q, and Joint under the paired sharing and supervision controls, retaining all measured task endpoints.

\begin{table}[htbp]
\centering\footnotesize\setlength{\tabcolsep}{1pt}
\caption{Absolute controls (\%; mean $\pm$ SD, five paired seeds). Task-isolated retains within-task federation; Label Q removes Private-to-Q teaching; no return removes Q-to-Private KD and fusion. CIFAR/AG: accuracy; I2T/T2I: Flickr R@1. Dashes: unreported.}
\label{tab:mechanism-absolute}\label{tab:followup-endpoints}

\begin{tabular*}{\columnwidth}{@{\extracolsep{\fill}}lcccc@{}}
\toprule
Endpoint & CIFAR & AG & I2T & T2I \\
\midrule
\multicolumn{5}{@{}l}{\emph{Full}} \\
Private & $45.13\!\pm\!0.39$ & $49.73\!\pm\!0.48$ & $35.09\!\pm\!0.26$ & $30.38\!\pm\!0.45$ \\
Q & $50.32\!\pm\!0.48$ & $75.85\!\pm\!1.11$ & $37.86\!\pm\!0.82$ & $31.13\!\pm\!0.73$ \\
Joint & $53.59\!\pm\!1.28$ & $78.50\!\pm\!0.42$ & $42.37\!\pm\!0.51$ & $35.46\!\pm\!0.48$ \\
\midrule
\addlinespace[2pt]
\multicolumn{5}{@{}l}{\emph{Local}} \\
Private & $38.55\!\pm\!0.13$ & -- & $18.22\!\pm\!0.74$ & $15.08\!\pm\!0.26$ \\
\addlinespace[2pt]
\multicolumn{5}{@{}l}{\emph{No return}} \\
Private & $38.55\!\pm\!0.13$ & -- & $18.22\!\pm\!0.74$ & $15.08\!\pm\!0.26$ \\
Q & $48.63\!\pm\!0.30$ & -- & $34.80\!\pm\!0.85$ & $29.39\!\pm\!0.38$ \\
Joint & $45.62\!\pm\!0.95$ & -- & $37.41\!\pm\!0.58$ & $31.39\!\pm\!0.28$ \\
\addlinespace[2pt]
\multicolumn{5}{@{}l}{\emph{Task-isolated}} \\
Private & $41.66\!\pm\!0.22$ & -- & $33.57\!\pm\!0.32$ & $28.94\!\pm\!0.17$ \\
Q & $48.90\!\pm\!0.30$ & -- & $35.80\!\pm\!0.76$ & $29.79\!\pm\!0.56$ \\
Joint & $49.38\!\pm\!0.79$ & -- & $40.56\!\pm\!0.50$ & $34.38\!\pm\!0.24$ \\
\addlinespace[2pt]
\multicolumn{5}{@{}l}{\emph{Label Q}} \\
Private & $43.82\!\pm\!0.19$ & -- & $33.98\!\pm\!0.33$ & $29.19\!\pm\!0.24$ \\
Q & $48.91\!\pm\!0.31$ & -- & $35.22\!\pm\!0.84$ & $29.06\!\pm\!0.37$ \\
Joint & $52.26\!\pm\!0.98$ & -- & $40.47\!\pm\!0.87$ & $34.12\!\pm\!0.27$ \\
\bottomrule
\end{tabular*}
\end{table}

\section{KD and Fusion Effect Definitions}
\Cref{tab:kd-fusion-absolute} reports the complete factorial in the main paper. The definitions below explain its paired main effects and interactions.

\paragraph{Factorial identification.}
Let $Y_{ab,s}$ be performance in seed $s$, with Q$\rightarrow$Private KD enabled by $a\in\{0,1\}$ and fusion training by $b\in\{0,1\}$. The four configurations use neither term, KD alone, fusion alone, or both (Full). All retain Private$\rightarrow$Q teaching. We evaluate Private and Joint after each configuration. Let $\Delta_{\mathrm{KD},s}$ and $\Delta_{\mathrm{Fus},s}$ denote the balanced within-seed KD and fusion effects:
\begin{equation}
\Delta_{\mathrm{KD},s}=\tfrac12[(Y_{10,s}-Y_{00,s})+(Y_{11,s}-Y_{01,s})]
\end{equation}
and analogously $\Delta_{\mathrm{Fus},s}=\tfrac12[(Y_{01,s}-Y_{00,s})+(Y_{11,s}-Y_{10,s})]$. The within-seed interaction $I_s$ is
\begin{equation}
I_s=(Y_{11,s}-Y_{10,s})-(Y_{01,s}-Y_{00,s}).
\end{equation}
The balanced main effects decompose the Full-minus-Neither gain, $Y_{11,s}-Y_{00,s}$. Interaction quantifies how enabling KD changes fusion's contribution: positive $I_s$ denotes a larger contribution with KD. \Cref{tab:kd-fusion-effects,tab:kd-fusion-absolute} report means and sample SDs of the paired seed-level effects.

\paragraph{Coordination of KD and fusion.}
On CIFAR Private, KD increases fusion's contribution from $0.01\pp$ to $3.27\pp$, giving a $3.26\pp$ interaction. On AG Private, Full reaches $49.73\%$ versus $48.79\%$ for KD alone. On AG Joint, KD increases fusion's contribution from $0.88\pp$ to $4.36\pp$, giving a $3.48\pp$ interaction. These examples illustrate coordination in private-model training and Joint deployment. Illustrative differences use rounded means; tabulated paired effects use seed-level values.

\section{Capacity Controls and Inference Costs}
\label{sec:ensemble-controls}
Independent matched trains separate Private predictors within Joint's total parameter budget. Independent $2\times$ full uses two full-sized Private predictors. Dual checkpoint combines two states of one training run. Conditional Private denotes a private model trained with Q supervision and deployed on its own. Conditional Joint combines that model with Q at inference. Flickr Independent Q-trained combines two separately trained, Q-supervised Private predictors. UCF donors retain their multimodal predictor; extra branches are used for recipients. Mixture weights and second checkpoints are selected on DEV and frozen before TEST. In the ensemble rows, DEV $w$ is the weight of the second predictor: $w=0$ keeps the first and $w=1$ keeps the second. Different seeds can select different weights, so the table reports their mean and sample SD. This is separate from Q's internal learned gate.

\Cref{tab:ensemble-controls-full} reports prediction, inference costs, and DEV weights together. Profiling uses RTX 4090 and batch size 64; latency covers inference execution, excluding data loading and host--device transfer. GFLOPs denotes billions of floating-point operations, Params (M) millions of parameters, and MiB mebibytes. Reported FLOPs cover profiler-supported operators; 0.00 denotes rounding. Peak memory denotes measured incremental activation memory. Training and communication costs are accounted for separately.

\begin{table*}[tbp]
\centering\footnotesize\setlength{\tabcolsep}{1.3pt}
\caption{Capacity controls: prediction, inference cost, DEV weights, and Joint-minus-control contrasts. Scores/weights are mean $\pm$ SD over five seeds. UCF: accuracy/macro F1; Flickr: I2T/T2I R@1. Flickr DEV weights are stacked I2T/T2I. $\Delta$ uses UCF accuracy or mean bidirectional Flickr R@1. Applicable exact paired $p=0.0625$. $^{\dagger}$Flickr ensembles use cyclic combinations of reused replicas; these gains are descriptive, with independent-pair CI/$p$ omitted. UCF dual-checkpoint costs profile two branches; DEV selects the first for prediction.}
\label{tab:ensemble-controls-full}\label{tab:ensemble-controls-paired}\label{tab:ensemble-controls-resources}\label{tab:ensemble-controls-weights}\label{tab:followup-capacity}\label{tab:followup-resources}

\begin{tabular*}{\textwidth}{@{\extracolsep{\fill}}lcccccccc@{}}
\toprule
Predictor / control & Acc./I2T & F1/T2I & Params (M) & GFLOPs & ms/sample & Peak MiB & DEV $w$ & \shortstack{$\Delta$ (pp)\\$[95\%\ \mathrm{CI}]$} \\
\midrule
\multicolumn{9}{@{}l}{\emph{UCF-51: accuracy / macro F1}} \\
Single Private & $71.70\pm0.13$ & $70.99\pm0.22$ & 0.43 & 0.00 & 0.007 & 2.0 & -- & -- \\
Conditional Private & \second{$72.50\pm0.29$} & \second{$71.93\pm0.40$} & 0.43 & 0.00 & 0.007 & 2.0 & -- & \shortstack{$+10.33$\\$[9.18,11.48]$} \\
Independent matched & $71.77\pm0.17$ & $71.04\pm0.32$ & 0.60 & 0.01 & 0.012 & 2.0 & $0.70\pm0.45$ & \shortstack{$+11.06$\\$[9.99,12.13]$} \\
Independent $2\times$ full & $71.70\pm0.19$ & $70.99\pm0.28$ & 0.86 & 0.01 & 0.014 & 2.0 & $0.44\pm0.52$ & \shortstack{$+11.13$\\$[9.98,12.28]$} \\
Private dual checkpoint & $71.70\pm0.13$ & $70.99\pm0.22$ & 0.86 & 0.01 & 0.014 & 2.0 & $0.00\pm0.00$ & \shortstack{$+11.13$\\$[9.97,12.29]$} \\
Conditional Joint & \best{$82.83\pm0.93$} & \best{$80.95\pm1.25$} & 0.60 & 0.01 & 0.012 & 2.0 & -- & -- \\
\midrule
\multicolumn{9}{@{}l}{\emph{Flickr30k: I2T / T2I R@1}} \\
Conditional Private & $35.09\pm0.26$ & $30.38\pm0.45$ & 18.07 & 3.64 & 0.147 & 392.0 & -- & \shortstack{$+6.18$\\$[5.76,6.60]$} \\
Independent Q-trained & \second{$36.56\pm0.21$} & \second{$31.54\pm0.05$} & 36.15 & 7.29 & 0.292 & 392.1 & \shortstack{$0.53\pm0.15$\\$0.47\pm0.07$} & $+4.87^{\dagger}$ \\
Independent Local & $22.29\pm0.56$ & $18.89\pm0.29$ & -- & -- & -- & -- & \shortstack{$0.51\pm0.15$\\$0.48\pm0.10$} & $+18.33^{\dagger}$ \\
Private dual checkpoint & $21.58\pm1.03$ & $18.39\pm0.67$ & 36.15 & 7.29 & 0.293 & 392.1 & \shortstack{$0.57\pm0.14$\\$0.53\pm0.13$} & \shortstack{$+18.93$\\$[17.61,20.25]$} \\
Conditional Joint & \best{$42.37\pm0.51$} & \best{$35.46\pm0.48$} & 36.15 & 7.27 & 0.260 & 392.1 & -- & -- \\
\bottomrule
\end{tabular*}
\end{table*}

\section{Interpreting Training and Inference Gains}
\label{sec:identification}
\paragraph{Training and inference contrasts.}
For a fixed dataset, seed, population, and metric, write $L_s$, $P_s$, and $J_s$ for independently trained Local, Q-trained Private, and Joint performance. The exact accounting identity is
\begin{equation}
 J_s-L_s=(P_s-L_s)+(J_s-P_s).
\label{eq:gain-accounting}
\end{equation}
The first term measures what Private retains after Q-assisted training; the second measures what keeping Q adds at inference. Panel (c) of \cref{tab:recipient-endpoints} separates retained training gains from the contribution of Q at deployment.

\section{Implementation and Reproducibility}
\label{sec:reproducibility}
\paragraph{Interface and private-model details.}
\label{sec:implementation-details}
Same-task Q inputs are 300-dimensional precomputed text features for MELD and eight 1280-dimensional visual tokens for CREMA-D/UCF-51. Donor-only audio features have 1611 dimensions on MELD and 852 on AV datasets. MELD donors combine text and audio bottlenecks. CREMA-D/UCF donors combine attention-pooled video and audio through visual, audio, gated, and joint heads. Task-heterogeneous Q uses frozen ImageNet-pretrained ResNet-18 image features and GloVe-mean text features with fixed orthogonal projection, each yielding 256 dimensions. These fixed mappings are separate from trainable Private encoders and remain local during communication rounds.

\paragraph{Optimization and release.}
The task-heterogeneous configuration trains for 40 communication rounds after a 10-epoch private warm-up. Per-round local schedules use two image epochs, two text epochs, and five Flickr30k epochs; learning rates are $5\times10^{-2}$, $5\times10^{-4}$, $2\times10^{-4}$, and $3\times10^{-4}$ for image, text, multimodal-private, and Q updates, respectively. The KD, fusion, and relation weights are 0.5, 0.25, and 0.2. Development data select checkpoints and inference fusion weights; the classification training gate is learned locally (\cref{eq:classification-fusion-training}). Release contents and evaluation aggregation are specified in \cref{sec:release-manifest}.

\paragraph{Communication accounting.}
Let $P_Q$ be the number of transmitted Q parameters, $\mathcal K_r$ the participating clients in round $r$, $N_{\mathrm{round}}$ the total rounds, and $B_{\mathrm{meta}}$ the total metadata bytes. Float32 uses four bytes per parameter, giving uplink payload approximately $4P_Q\sum_{r=1}^{N_{\mathrm{round}}}|\mathcal K_r|+B_{\mathrm{meta}}$, where $|\mathcal K_r|$ counts participating clients. This excludes initial interface distribution, orchestration, storage, and transport framing. Each recorded task-heterogeneous run transfers 1.4609 decimal GB upstream and 1.4608 GB downstream, with a 3.652 MB Q state and 800 recorded messages.

\section{Evaluation Aggregation and Release}
\label{sec:release-manifest}
Same-task ``All'' accuracy pools test samples; heterogeneous classification averages fixed client endpoints; Flickr uses five-fold 200-image retrieval. Primary measured results use five seeds. The release will map each measured row to its seeds, evaluator, and experiment exports, and include code, configurations, split manifests, feature checkpoints, preprocessing, cross-fitting, and calibration.

\section{Backbone Groups}
\label{sec:followup}
\Cref{tab:followup-backbones} reports UCF deployment endpoints within each backbone's fixed client population. Accuracy, macro F1, and UAR characterize prediction across all three groups; paired Joint-minus-Private accuracy differences use the same clients within each group.

\begin{table}[htbp]
\centering\footnotesize\setlength{\tabcolsep}{1.5pt}
\caption{UCF backbone groups (\%; mean $\pm$ SD, five seeds). (a) Accuracy, macro F1, and UAR. (b) Paired Joint-minus-Private accuracy gains (pp), 95\% CIs, and positive-seed counts. Each group has a fixed test population; Trans. denotes Transformer.}
\label{tab:followup-backbones}\label{tab:backbones-main}\label{tab:followup-backbone-gains}

\textit{(a) Prediction endpoints}\par\smallskip
\begin{tabular*}{\columnwidth}{@{\extracolsep{\fill}}llccc@{}}
\toprule
Backbone & Mode & Acc. & F1 & UAR \\
\midrule
MLP & Private & $76.23\!\pm\!0.50$ & $67.60\!\pm\!0.82$ & $71.94\!\pm\!0.75$ \\
 & Q & $83.04\!\pm\!2.06$ & $76.79\!\pm\!2.55$ & $79.73\!\pm\!2.19$ \\
 & Joint & $91.25\!\pm\!0.98$ & $87.83\!\pm\!1.52$ & $87.77\!\pm\!1.25$ \\
\midrule
CNN & Private & $61.31\!\pm\!1.95$ & $55.71\!\pm\!2.48$ & $63.09\!\pm\!2.32$ \\
 & Q & $77.69\!\pm\!0.97$ & $71.51\!\pm\!0.64$ & $79.48\!\pm\!0.89$ \\
 & Joint & $79.88\!\pm\!0.97$ & $77.13\!\pm\!1.20$ & $82.93\!\pm\!0.71$ \\
\midrule
Trans. & Private & $80.05\!\pm\!1.53$ & $69.85\!\pm\!1.41$ & $72.75\!\pm\!1.49$ \\
 & Q & $75.67\!\pm\!1.75$ & $74.38\!\pm\!2.53$ & $78.74\!\pm\!1.87$ \\
 & Joint & $81.87\!\pm\!2.97$ & $80.77\!\pm\!3.03$ & $83.43\!\pm\!2.35$ \\
\bottomrule
\end{tabular*}
\par\smallskip
\textit{(b) Joint-minus-Private accuracy (pp)}\par\smallskip
\begin{tabular*}{\columnwidth}{@{\extracolsep{\fill}}lccc@{}}
\toprule
Backbone & Gain & 95\% CI & Positive \\
\midrule
MLP & $+15.02\!\pm\!1.41$ & $[+13.27,+16.77]$ & 5/5 \\
CNN & $+18.57\!\pm\!2.30$ & $[+15.71,+21.42]$ & 5/5 \\
Trans. & $+1.82\!\pm\!2.99$ & $[-1.90,+5.53]$ & 3/5 \\
\bottomrule
\end{tabular*}
\end{table}

\begin{table*}[tbp]
\centering\footnotesize\setlength{\tabcolsep}{3pt}
\caption{Direct shared-model endpoints (\%; mean $\pm$ SD, five seeds). F1 is macro F1; Flickr uses five-fold R@1. Group headers give shared parameters, 40-round uplink, and aggregation. FedSocket uses primary endpoints. Paired Joint gains appear in \cref{tab:neighbor-paired}.}
\label{tab:neighbor-endpoints}\label{tab:neighbor-resources}
\footnotesize\setlength{\tabcolsep}{2pt}\renewcommand{\arraystretch}{1.05}

\begin{tabular*}{\textwidth}{@{\extracolsep{\fill}}lcccccc@{}}
\toprule
Endpoint & CIFAR Acc. & CIFAR F1 & AG Acc. & AG F1 & I2T R@1 & T2I R@1 \\
\midrule
\multicolumn{7}{@{}l}{\emph{FedSocket}: 913,003 shared parameters; 1.4609 GB uplink; task-owned / conflict-aware} \\
Private & $45.13\!\pm\!0.39$ & $36.88\!\pm\!0.43$ & $49.73\!\pm\!0.48$ & $38.79\!\pm\!0.65$ & $35.09\!\pm\!0.26$ & $30.38\!\pm\!0.45$ \\
Shared Q & $50.32\!\pm\!0.48$ & $49.44\!\pm\!0.48$ & $75.85\!\pm\!1.11$ & $74.25\!\pm\!1.57$ & $37.86\!\pm\!0.82$ & $31.13\!\pm\!0.73$ \\
Joint & $53.59\!\pm\!1.28$ & $49.70\!\pm\!1.23$ & $78.50\!\pm\!0.42$ & $76.90\!\pm\!0.42$ & $42.37\!\pm\!0.51$ & $35.46\!\pm\!0.48$ \\
\midrule
\multicolumn{7}{@{}l}{\emph{FML-adapt.}: 913,052 shared parameters; 1.4609 GB uplink; equal-client mean} \\
Private & $38.40\!\pm\!0.14$ & $27.04\!\pm\!0.16$ & $48.00\!\pm\!0.14$ & $35.69\!\pm\!0.21$ & $26.29\!\pm\!0.32$ & $21.87\!\pm\!0.38$ \\
Shared Q & $35.03\!\pm\!0.16$ & $30.20\!\pm\!0.21$ & $74.38\!\pm\!0.79$ & $72.72\!\pm\!0.77$ & $35.20\!\pm\!0.53$ & $28.86\!\pm\!0.77$ \\
Joint & $41.17\!\pm\!0.49$ & $31.79\!\pm\!0.70$ & $74.63\!\pm\!0.79$ & $73.19\!\pm\!0.86$ & $38.62\!\pm\!0.65$ & $32.31\!\pm\!0.55$ \\
\midrule
\multicolumn{7}{@{}l}{\emph{FedKD-adapt.}: 913,052 shared parameters; $0.9576\pm0.0019$ GB uplink; sample-weighted / SVD} \\
Private & $38.27\!\pm\!0.11$ & $26.90\!\pm\!0.10$ & $48.42\!\pm\!0.17$ & $36.19\!\pm\!0.23$ & $24.24\!\pm\!0.73$ & $19.90\!\pm\!0.32$ \\
Shared Q & $33.51\!\pm\!0.33$ & $28.67\!\pm\!0.47$ & $69.17\!\pm\!1.89$ & $63.68\!\pm\!2.76$ & $36.28\!\pm\!0.94$ & $29.88\!\pm\!0.60$ \\
Joint & $40.10\!\pm\!0.19$ & $30.14\!\pm\!0.32$ & $70.27\!\pm\!1.19$ & $67.84\!\pm\!1.29$ & $38.54\!\pm\!0.54$ & $32.23\!\pm\!0.54$ \\
\midrule
\multicolumn{7}{@{}l}{\emph{Naive Shared}: 913,052 shared parameters; 1.4609 GB uplink; sample-weighted mean} \\
Private & $37.26\!\pm\!0.09$ & $26.10\!\pm\!0.10$ & $48.26\!\pm\!0.18$ & $36.14\!\pm\!0.23$ & $35.24\!\pm\!0.39$ & $29.81\!\pm\!0.41$ \\
Shared Q & $33.45\!\pm\!0.42$ & $28.77\!\pm\!0.56$ & $68.56\!\pm\!1.01$ & $61.75\!\pm\!1.58$ & $37.68\!\pm\!0.79$ & $30.62\!\pm\!0.48$ \\
Joint & $39.35\!\pm\!0.38$ & $29.75\!\pm\!0.65$ & $70.14\!\pm\!0.73$ & $66.97\!\pm\!0.77$ & $41.07\!\pm\!0.64$ & $34.48\!\pm\!0.22$ \\
\bottomrule
\end{tabular*}

\end{table*}

\begin{table}[htbp]
\centering\footnotesize\setlength{\tabcolsep}{1pt}
\caption{Paired Joint gains over shared-model controls (pp; five seeds). Each cell gives mean $\pm$ SD and unadjusted 95\% $t$ CI. The exact two-sided sign-flip $p=0.0625$ applies to a five-of-five positive sign pattern. CIFAR/AG: accuracy; Flickr: R@1.}
\label{tab:neighbor-paired}
\begin{tabular*}{\columnwidth}{@{\extracolsep{\fill}}lccc@{}}
\toprule
Metric & FML-adapt. & FedKD-adapt. & Naive Shared \\
\midrule
CIFAR Acc. & \shortstack{$+12.42\pm1.20$\\$[+10.93,+13.91]$} & \shortstack{$+13.49\pm1.31$\\$[+11.86,+15.12]$} & \shortstack{$+14.24\pm1.10$\\$[+12.88,+15.60]$} \\
\addlinespace[2pt]
AG Acc. & \shortstack{$+3.87\pm0.80$\\$[+2.88,+4.86]$} & \shortstack{$+8.23\pm1.18$\\$[+6.77,+9.69]$} & \shortstack{$+8.36\pm0.70$\\$[+7.49,+9.23]$} \\
\addlinespace[2pt]
I2T R@1 & \shortstack{$+3.74\pm0.81$\\$[+2.74,+4.74]$} & \shortstack{$+3.83\pm0.45$\\$[+3.27,+4.38]$} & \shortstack{$+1.29\pm0.84$\\$[+0.25,+2.33]$} \\
\addlinespace[2pt]
T2I R@1 & \shortstack{$+3.15\pm0.70$\\$[+2.29,+4.02]$} & \shortstack{$+3.23\pm0.58$\\$[+2.51,+3.96]$} & \shortstack{$+0.98\pm0.58$\\$[+0.26,+1.70]$} \\
\bottomrule
\end{tabular*}
\end{table}

\section{Direct Comparisons with Shared-Model Neighbors}
\label{sec:neighbor-controls}
\paragraph{Common protocol.}
All methods use the same 3/3/4 CIFAR/AG/Flickr clients, fixed partitions, private backbones, 256-dimensional interfaces, local schedules, 40 rounds, DEV checkpoint/fusion selection, and TEST evaluator.

\paragraph{Baseline adaptations.}
Five seeds ($101,202,303,404,505$) vary training randomness on one partition. TEST is evaluated once per method/seed with the same Joint post-processing. FML-adapted~\cite{shen2020fml} uses mutual Private/shared learning and equal-client averaging. FedKD-adapted~\cite{wu2022fedkd} uses Private mentors, a shared mentee, reliability-weighted output/hidden distillation, singular-value decomposition (SVD)-compressed large-matrix updates, and sample-weighted aggregation. Naive Shared uses a monolithic shared backbone, bidirectional distillation, and sample-weighted averaging. The three controls use the same architecture and initialization mapped from one Flickr shared checkpoint. They have 913,052 parameters versus Q's 913,003, a difference below 0.006\%. Their outputs are adapted to this benchmark.

\paragraph{DEV selection.}
Each baseline tests three candidates for 10 DEV-only rounds on seed 101, then fixes its hyperparameters for all five runs. FML task-loss weights are $(\alpha_{\mathrm{FML}},\beta_{\mathrm{FML}})\in\{(0.5,0.5),(0.7,0.7),(0.9,0.9)\}$, with $(0.9,0.9)$ selected. FedKD tests hidden-KD weights $\{0,0.1,0.2\}$ and selects $0.2$. Naive Shared tests KD weights $\{0.25,0.5,0.75\}$ and selects $0.75$. Checkpoints and mixtures are selected on DEV and fixed before TEST.

\Cref{tab:neighbor-endpoints,tab:neighbor-paired} report deployment endpoints, shared parameters, communication budgets, and paired Joint contrasts. Upload is cumulative upstream traffic over 40 rounds in decimal GB. FedSocket, FML-adapted, and Naive Shared share a 1.4609 GB uplink budget; FedKD's compression reduces upload by 34.45\%, providing a compressed operating point for the accuracy--communication comparison.

\begin{figure*}[tbp]
\centering
\includegraphics[width=\linewidth]{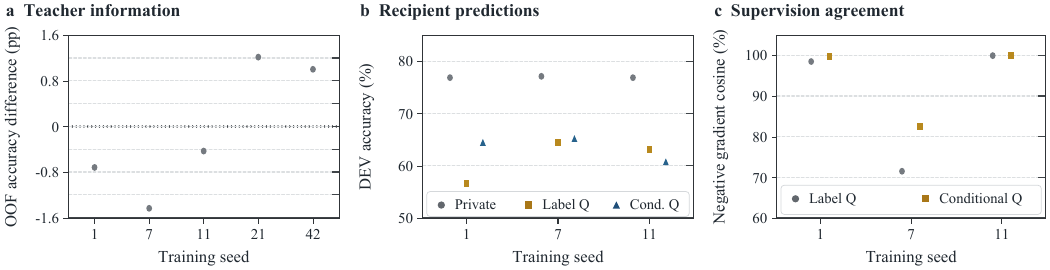}
\caption{Teacher information and recipient compatibility (D01). (a) Full-modality minus video-only teacher accuracy on 1,395 out-of-fold (OOF) donor examples, in percentage points (pp). (b) Historical predictions on the same 813 recipient DEV examples: Private is the local model; Label Q is label-trained Q; Cond. Q is Conditional Q trained with full-modality teacher supervision. (c) Percentage of correct-teacher TRAIN observations with negative cosine between cross-entropy (CE) and knowledge-distillation (KD) gradients at the private logits. Each point is an observed seed-level value; panels (b,c) use seeds 1, 7, and 11.}
\label{fig:diag-d01}
\end{figure*}

\begin{figure*}[tbp]
\centering
\includegraphics[width=\linewidth]{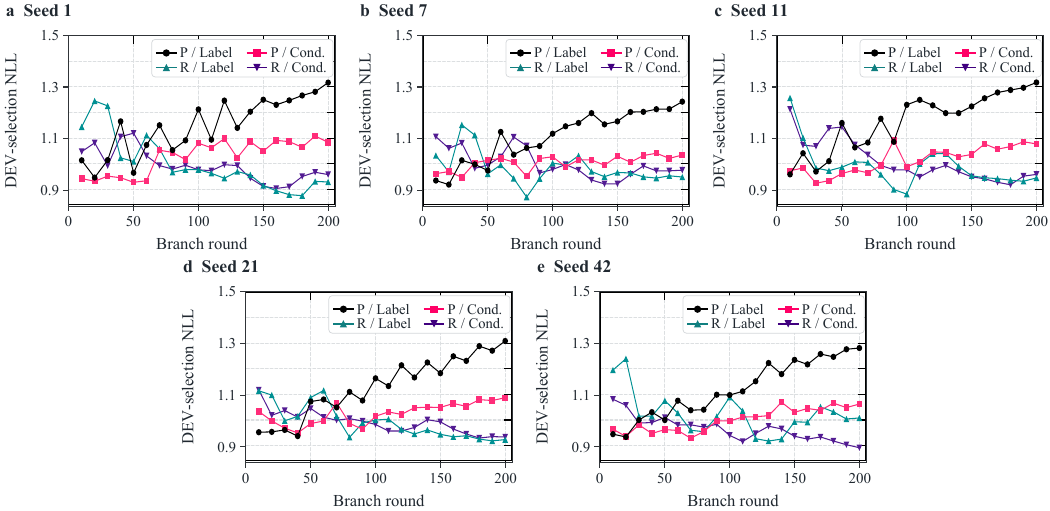}
\caption{Q optimization on the 383-example DEV-selection split (D02); lower negative log-likelihood (NLL) is better. Each panel shows one seed and 20 observations per curve. P/R: persistent/reset server Adam state; Label/Cond.: label-only/full-modality-teacher supervision. Branch round is recorded round minus 100. These curves are DEV-selection candidates, including the legacy optimizer diagnostic.}
\label{fig:diag-d02}
\end{figure*}

\begin{figure*}[tbp]
\centering
\includegraphics[width=\linewidth]{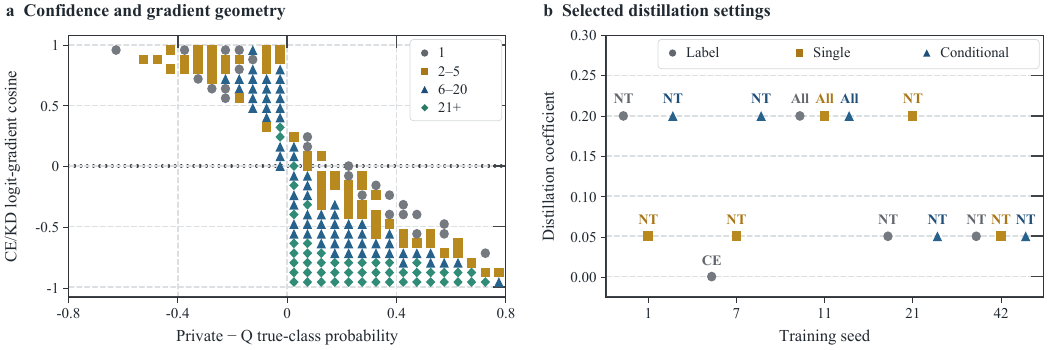}
\caption{Returned supervision and selected distillation (D04). (a) 6,096 correct-teacher Conditional-Q TRAIN observations (seeds 1, 7, 11), binned on a $32\times25$ grid with 198 occupied bins. Axes: Private minus Q true-class probability and CE/KD logit-gradient cosine. Symbols: observations per bin (1, 2--5, 6--20, 21+). (b) DEV-selected KD coefficients and modes: NT is non-target-class KL; All is all-class KL; CE uses labels without KD (coefficient zero). Joint mixture weights are selected separately.}
\label{fig:diag-d04}
\end{figure*}

\section{Recipient-Side Development Diagnostics}
\label{sec:development-diagnostics}
\begingroup
\captionsetup{skip=4pt}
These CREMA-D diagnostics use 27 multimodal donors, 64 video-only recipients, fixed features and client assignments, a 214,663-parameter Q, and seeds $1,7,11,21,42$. Historical recipient development (DEV) data contain 813 examples, group-split into 383 selection and 430 audit examples. The audit is a holdback from previously observed DEV data. \Cref{fig:diag-d01,fig:diag-d02,fig:diag-d03,fig:diag-d04} report OOF, TRAIN, and DEV measurements; primary TEST and Set A results retain their original protocols.

\par\smallskip\noindent\textbf{Teacher information and recipient compatibility.} 
\Cref{fig:diag-d01} separates donor teacher quality from prediction through recipient-observed inputs. The gradient diagnostic examines how transferred supervision interacts with local labels.

\par\smallskip\noindent\textbf{Selecting useful Q checkpoints.} 
Q receives 100 supervised warm-up rounds and a 200-round branch budget. Recipient DEV negative log-likelihood (NLL) selects the checkpoint; \cref{fig:diag-d02} shows its evolution across rounds and supervision sources.

\par\smallskip\noindent\textbf{Evaluating Q as a prediction partner.} 
The two-stage development protocol first selects Q, then trains recipient Private models and selects their joint prediction rule. Private receives 20 CE warm-up epochs and up to 180 further epochs, with selection patience 30. Candidates use all-class or non-target-class Kullback--Leibler (KL) distillation with coefficient $0.05$ or $0.2$, together with a CE-only candidate. Only correct-teacher TRAIN examples activate KD. Private selection maximizes DEV accuracy, using NLL to break ties; probability-mixture weights are selected from $\{0,0.25,0.5,0.75,1\}$. All choices are frozen before development-audit inference.

Conditional Q attains the highest mean Joint accuracy among these development candidates, $76.60\pm0.42\%$, versus its Private endpoint at $76.28\pm0.72\%$ and Q at $69.44\pm0.92\%$ (\cref{fig:diag-d03}). The paired matrix shows seed-level Conditional-minus-Label changes.

\begin{samepage}
\par\smallskip\noindent\textbf{Connecting supervision to the recipient's prediction.} 
\Cref{fig:diag-d04}(a) expands the supervision diagnostic in \cref{fig:recipient-supervision}; panel (b) gives the selected return-path settings. Non-target KL transfers relations among alternative classes. DEV selection chooses the mode and strength for each source and seed within this two-stage protocol.

\end{samepage}
\endgroup
\end{document}